\documentclass[10pt,dvipsnames]{article} % For LaTeX2e
\usepackage[accepted]{tmlr}
\usepackage[utf8]{inputenc} % allow utf-8 input
\usepackage[T1]{fontenc}    % use 8-bit T1 fonts
\usepackage{hyperref}       % hyperlinks
\usepackage{url}            % simple URL typesetting
\usepackage{booktabs}       % professional-quality tables
\usepackage{amsfonts}       % blackboard math symbols
\usepackage{nicefrac}       % compact symbols for 1/2, etc.
\usepackage{microtype}      % microtypography
\usepackage{amsmath}
\usepackage{amssymb}
\usepackage{mathtools}
\usepackage{amsthm}
\usepackage{algorithm}
\usepackage{algpseudocode}
\usepackage[capitalize,noabbrev]{cleveref}
\usepackage{bm}
\usepackage{array}
\usepackage{subcaption}
\usepackage{multirow}

\newcommand{\algName}[0]{NTD-CFE}
\newcommand{\fullAlgName}[0]{No-Training-Dataset reinforcement-learning-based CFE}

\newcommand{\MaxEpisodes}[0]{M_E}
\newcommand{\MaxInterventions}[0]{M_T}
\newcommand{\RLParameter}[0]{\theta}
\newcommand{\RLParameters}[0]{\bm{\RLParameter}}
\newcommand{\policyAgent}[0]{\pi_{{\RLParameters}}}
\newcommand{\policy}[2]{\policyAgent(#1|#2)}
\newcommand{\PredModel}[0]{f}
\newcommand{\RewardFunc}[0]{R}
\newcommand{\RLState}[0]{s}
\newcommand{\Reward}[0]{r}
\newcommand{\RLReturn}[0]{G}
\newcommand{\DiscountFactor}[0]{\gamma}
\newcommand{\LearningRate}[0]{\alpha}
\newcommand{\WeightDecay}[0]{\lambda_{\textit{WD}}}
\newcommand{\StateTransFunc}[0]{F_{p}}
\newcommand{\ProxmtDistance}[0]{D_{\textit{pxmt}}}

\newcommand{\InDistrib}[0]{\text{in\_dist}}
\newcommand{\InDistribFunc}[0]{F_{\InDistrib}}

\newcommand{\ProxmtLambda}[0]{\lambda_{\textit{pxmt}}}

\newcommand{\FeasbltWeights}[0]{W_{\textit{fsib}}}

\newcommand{\CausalConstrants}[0]{C_{\textit{causal}}}
\newcommand{\RangeConstrants}[0]{C_{\textit{range}}}

\newcommand{\userInput}[0]{\bm{x}^*}
\newcommand{\targetClass}[0]{Y'}

\newcommand{\action}[0]{\bm{a}}
\newcommand{\actionDimensionOne}[0]{a_{\text{time}}}
\newcommand{\actionDimensionTwo}[0]{a_{\text{feat}}}
\newcommand{\actionDimensionThree}[0]{a_{\text{stre}}}

\newcommand{\actionDimensionOneParameters}[0]{\bm{p_{\text{time}}}}
\newcommand{\actionDimensionTwoParameters}[0]{\bm{p_{\text{feat}}}}
\newcommand{\actionDimensionThreeParametersMean}[0]{\bm{\mu_{\text{DC}}}}
\newcommand{\actionDimensionThreeParameterMean}[1]{\mu_{\text{DC}, {#1}}}
\newcommand{\actionDimensionThreeParametersSTD}[0]{\bm{\sigma_{\text{DC}}}}
\newcommand{\actionDimensionThreeParameterSTD}[1]{\sigma_{\text{DC}, {#1}}}
\newcommand{\actionDimensionThreeParametersDiscreteValue}[0]{\bm{p_{\NumValuesDiscFeat}}}
\newcommand{\actionDimensionThreeParameterDiscreteValue}[1]{\bm{p_{\NumValuesDiscFeat, {#1}}}}

\newcommand{\NumValuesDiscFeat}[0]{N_{\text{dis}}}
\newcommand{\NumValuesDiscFeatIth}[1]{N_{\text{dis}, {#1}}}
\newcommand{\discreteFeatures}[0]{\bm{D_{\text{dis}}}}
\newcommand{\actFeatures}[0]{\bm{D_{\text{act}}}}
\newcommand{\nonActFeatures}[0]{\bm{D_{\text{non-act}}}}
\newcommand{\immuFeatures}[0]{\bm{D_{\text{immu}}}}

\newcommand{\NumInvalidSamples}[0]{N_{\textit{inv}}}

\newcommand{\rlCodeColor}[0]{gray}
\newcommand{\resultIgnoreColor}[0]{gray}
\newcommand{\codeCommentColor}[0]{teal}

\title{No $D_{\text{train}}$: Model-Agnostic Counterfactual Explanations Using Reinforcement Learning}

\author{\name Xiangyu Sun \email xiangyu.sun@rbcborealis.com \\
\addr RBC Borealis
\AND
\name Raquel Aoki \email raquel.aoki@rbcborealis.com \\
\addr RBC Borealis
\AND
\name Kevin H. Wilson \email kevin.h.wilson@rbcborealis.com\\
\addr RBC Borealis}

\def\month{06}  % Insert correct month for camera-ready version
\def\year{2025} % Insert correct year for camera-ready version
\def\openreview{\url{https://openreview.net/forum?id=egNzAG9rOu}} % Insert correct link to OpenReview for camera-ready version

\begin{document}

\maketitle

\begin{abstract}
Machine learning (ML) methods have experienced significant growth in the past decade, yet their practical application in high-impact real-world domains has been hindered by their opacity. When ML methods are responsible for making critical decisions, stakeholders often require insights into how to alter these decisions. Counterfactual explanations (CFEs) have emerged as a solution, offering interpretations of opaque ML models and providing a pathway to transition from one decision to another. However, most existing CFE methods require access to the model's training dataset, few methods can handle multivariate time-series, and none of model-agnostic CFE methods can handle multivariate time-series without training datasets. These limitations can be formidable in many scenarios. In this paper, we present \algName{}, a novel model-agnostic CFE method based on reinforcement learning (RL) that generates CFEs when training datasets are unavailable. \algName{} is suitable for both static and multivariate time-series datasets with continuous and discrete features. \algName{} reduces the CFE search space from a multivariate time-series domain to a lower dimensional space and addresses the problem using RL. Users have the flexibility to specify non-actionable, immutable, and preferred features, as well as causal constraints. We demonstrate the performance of \algName{} against four baselines on several datasets and find that, despite not having access to a training dataset, \algName{} finds CFEs that make significantly fewer and significantly smaller changes to the input time-series. These properties make CFEs more actionable, as the magnitude of change required to alter an outcome is vastly reduced. The code is available in the supplementary material.
\end{abstract}

\section{Introduction}

After receiving a negative decision---a denial of a loan, a poor performance review, or a rejection from a prestigious conference---a very natural question to ask is ``What could I have done differently?'' When the decision is made by a person, that question can be answered directly by the person. Indeed, peer reviews represent the reasons why a paper is accepted or rejected from a conference and, in the best case, give authors actionable feedback which may lead to an acceptance in the future (assuming the underlying decision algorithm remains unchanged). But when a decision is made or influenced by a black-box model, it can be much harder to provide insights. Telling a loan applicant they have a ``poor credit score'' does not tell them how they might approach getting approved at a later date.

Counterfactual Explanations (CFEs) \citep{wachter2017counterfactual} were introduced to fill this gap. Given an input to a model, a CFE is perturbed version of the input which yields a prescribed output from the model. For instance, if Bob's mortgage application is rejected, a CFE might suggest that Bob make \$20,000 more per year, or alternatively, make \$10,000 more per year and purchase a house in a different neighborhood.

CFEs also provide a method for examining how ``fair'' a model is, a concern that has become paramount in many real-world applications \citep{mehrabi2021survey, angwin2022machine, osoba2017intelligence}. For instance, a CFE that shows if Bob's name were Alice his loan application would have been accepted points to potential discrimination on the basis of a protected characteristic. Classic models such as logistic regression and low-depth decision trees often called inherently interpretable \citep{survey_cfe_verma2020counterfactual, murphy2012machine, breiman2017classification} as the relative influence of input features can be read off from learned coefficients. But relations between learned parameters and input features are much harder to discern with more complex models. Indeed, even a logistic regression model with pairwise interaction terms can be complicated to reason about: Perhaps a woman who purchases a home in one neighborhood is less likely to have her application accepted than a man, but an additional \$1,000 a year in income increases her chance of acceptance significantly more than a man's chance.

There exist many CFE methods for static datasets \citep{dice_mothilal2020explaining, cfrl_samoilescu2021model, fastar_verma2022amortized}. However, CFE methods for multivariate time-series data are less common due to the challenges posed by higher dimensionality \citep{comte_ates2021counterfactual,theissler2022explainable}. Additionally, to the best of our knowledge, {\em all existing model-agnostic CFE methods for multivariate time-series require access to large collection of samples from the training distribution of the model being explained}. This requirement can be infeasible in real-world domains especially due to privacy and other concerns.

In this paper, we introduce \fullAlgName{} (\algName{}), a reinforcement learning (RL) based CFE method designed for both static and multivariate time-series data containing both continuous and discrete attributes. Remarkably, \algName{} operates without training datasets or similar data samples and is model-agnostic so that it is compatible with any (even non-differentiable) predictive models. \algName{} also allows the user to specify which features they prefer to change, as well as how changing a particular feature may affect another feature, thus allowing the user to express both what counterfactuals are feasible for them and any causal relationships between those features. While \algName{} works for both static and time-series data, we focus on the harder setting of multivariate time-series data in this paper. We compare \algName{} to four state-of-the-art CFE methods on nine real-world multivariate time-series datasets. We find that \algName{} yields CFEs with significantly better proximity (the total magnitude of change proposed by the CFE) and sparsity (how many features the CFE proposes to alter) compared to the baselines.

The paper is structured as follows. We discuss related works and preliminaries in~\Cref{section_related_works} and~\Cref{section_preliminaries}, respectively. \Cref{section_proposed_algorithm} describes the proposed algorithm \algName{}. 
% Qualitative examples and quantitative 
Experiments with nine real-world datasets and five predictive models are given in~\Cref{section_evaluation}. 
Finally, we conclude in~\Cref{section_conclusion}.

\section{Related Works}\label{section_related_works}

\paragraph{Explainable AI (XAI)} Counterfactual explanations belong to a much broader category of methods often called Explainable AI. XAI techniques can be broadly classified into two buckets \citep{survey_cfe_verma2020counterfactual}: (a) inherently interpretable machine learning techniques and (b) post-hoc explanatory techniques. The first category is primarily a restriction on the types of models that a practitioner can employ to model a phenomenon. However, models often held out as interpretable (e.g., linear regression and low-depth decision trees \citep{murphy2012machine, breiman2017classification}) may not have the capacity to capture complex phenomena. In the latter category, practitioners attempt to ``model the model'' with subsequent techniques \citep{sun2020cracking,liu2021learning}. These methods can be further subdivided into global and local methods \citep{comte_ates2021counterfactual, islam2021explainable}. Global methods attempt to simulate the opaque, complex logic of the original model using interpretable methods. On the other hand, local methods aim to explain the rationale behind a specific prediction made for a specific input. For instance, feature-based methods such as SHAP\citep{lundberg2017unified}, identify features that have the most significant influence on a prediction. On the other hand, sample-based methods attempt to identify relevant samples to clarify a prediction \citep{kim2016examples, dice_mothilal2020explaining}. CFEs are an example of a post-hoc, local, sample-based explanation technique.

\paragraph{Counterfactual Explanations (CFEs)} CFEs were introduced by \citet{wachter2017counterfactual} to explore an optimization-based technique for differentiable predictive models. Building upon this foundation, DiCE \citep{dice_mothilal2020explaining} noted that there were potentially many different CFEs proximal to any particular input and whether one was ``better'' than another was really a matter for the CFE's user to decide. As such, DiCE returns several diverse CFEs for any given sample. However, the base algorithm of DiCE is not guaranteed to return CFEs which satisfy causal constraints, and so DiCE introduced an expensive pruning step to filter ``non-feasible'' CFEs, i.e., those that do not satisfy causal constraints. 
% A method that directly incorporated these causal constraints into the exploration phase would likely improve the time to generate CFEs.
These optimization-based methods are, unfortunately, not model-agnostic, as they require the underlying predictive model to be differentiable and thus exclude popular predictive models such as random forests and $k$-nearest neighbors. To overcome this limitation, methods such as those of \citet{tsirtsis2021counterfactual} leverage dynamic programming to identify an optimal counterfactual policy and subsequently utilize this policy to find CFEs. However, due to the high memory requirements of these methods, this technique is best suited for low-dimensional data with discrete features. Another line of work employs Bayesian optimization (BO) to address the CFE task \citep{spooner2021counterfactual,romashov2022baycon,huang2024tx}. While these methods do not require training datasets, due to computational inefficiency of BO, they address static or univariate time-series data rather than multivariate time-series data.

\paragraph{RL-based Methods} CFRL \citep{cfrl_samoilescu2021model} and FastAR \citep{fastar_verma2022amortized} take a different tactic, employing techniques from RL to generate CFEs. CFRL first encodes samples into latent space using autoencoders \citep{kingma2013auto}, then an RL agent is trained to find a CFE in the latent space. Finally, a decoder converts the latent CFE back to the input space. Similarly, FastAR converts the CFE problem to a Markov decision process (MDP) \citep{sutton2018reinforcement} and then uses proximal policy optimization (PPO) \citep{schulman2017proximal} to solve the MDP. However, the action candidates in FastAR are discrete and fixed, making it impractical for complex multivariate time-series data. Moreover, both CFRL and FastAR require access to training datasets.

\paragraph{CFEs for Time-series Data} These and several other CFE techniques cater to static data, but methods for handling multivariate time-series data are much less prevalent \citep{theissler2022explainable}. For time-series data with $k$-nearest neighbor or random shapelet forest models, \citet{karlsson2020locally} introduce one approach. \citet{wang2021learning} focus on univariate time-series, seeking CFEs in a latent space before decoding them back to the input space. Native-Guide \citep{native_guide_delaney2021instance} finds the nearest unlike neighbor of an original univariate time-series sample, identifies the most influential subsequence of the neighbor, and substitutes it for the corresponding region in the original sample. On the other hand, CoMTE \citep{comte_ates2021counterfactual} can handle multivariate time-series data. First, it searches for distractor candidates, which are the original sample's neighbors in the training dataset that are predicted as the target class. Then, it identifies the best substitution parts on each distractor candidate. Finally, it gives a CFE by replacing a corresponding part of the original sample with the best substitution of the best distractor candidate.

\paragraph{Adversarial Learning} We also note that, on the surface, generating adversarial examples seems quite similar to generating CFEs: both create proximal samples that yield distinct predictions from the original inputs. However, while adversarial learning considers proximity, essential CFE properties such as actionability, feasibility, and causal constraints are mostly ignored \citep{wachter2017counterfactual, survey_cfe_verma2020counterfactual, sulem2022diverse}.

\section{Preliminaries}\label{section_preliminaries}

\paragraph{Problem Definition} CFEs aim to solve the following task: Given a user input $\userInput$, a predictive model $\PredModel$ and a target prediction $\targetClass$ such that $\PredModel(\userInput) \neq \targetClass$, the goal is to find a transformation from $\userInput$ to a new sample $\bm{x'}$ (i.e. CFE) such that~\citep{survey_cfe_verma2020counterfactual, karimi2020survey, guidotti2022counterfactual}:
\begin{itemize}
    \item valid: $\PredModel(\bm{x'}) = \targetClass$,
    \item actionable: if the user cannot modify feature $j$, then $\userInput_j = \bm{x'}_j$,
    \item sparse: the CFE $\bm{x'}$ should differ in as few features from $\userInput$ as possible,
    \item proximal: the distance between $\userInput$ and $\bm{x'}$ should be small in some metric, and
    \item plausible: $\bm{x'}$ should satisfy all causal constraints on the input.
\end{itemize}

%%% Incorporated into the above.

%, $\bm{x'}$ remains plausible (e.g., comes from the same distribution as $\userInput$), and is proximal (i.e., is close in some metric) to $\userInput$~\citep{survey_cfe_verma2020counterfactual, guidotti2022counterfactual}.

%CFEs may also require further desirable properties to be effective~\citep{survey_cfe_verma2020counterfactual, karimi2020survey, guidotti2022counterfactual}. A CFE $\bm{x'}$ should be valid: $\PredModel(\bm{x'}) = \targetClass$. The CFE should be actionable, i.e., it should only modify features which the user can change. Additionally, a generated CFE should demonstrate proximity to the original user input. This concept is closely related to the notion of \textit{feasibility}. A CFE is considered closer to the original user input if the change involves a more feasible feature as opposed to a less feasible feature. Feasibility also encodes a user's preference, since each user may prefer to change a different set of features. Moreover, a CFE should exhibit \textit{sparsity}, implying a minimal number of modified features. Furthermore, a CFE ought to be \textit{plausible}, with all features adhering to \textit{causal constraints} that exist among them. It ensures that the CFE represents an actual state of the world.

where $x_j$ denotes the $j$-th feature of $x$. 

\paragraph{Reinforcement Learning} Our method for finding CFEs will rely on reinforcement learning (RL). In RL, an agent and an environment interact with each other \citep{sutton2018reinforcement}. The agent takes an action $a_t$ on a state $s_t$ at time step $t$. The environment receives $a_t$ and $s_t$ from the agent and returns the next state $s_{t+1}$ and a reward $R_{t+1}$ to the agent. The goal of the agent is to maximize the expected cumulative (discounted) reward. RL can be categorized as model-free RL and model-based RL. In model-free RL \citep{mnih2015human, haarnoja2018soft}, the agent learns a policy from real experience when a model of the environment is not available to the agent. In model-based RL \citep{silver2017mastering,ha2018world}, the agent plans a policy from simulated experience generated by a model of the environment. This model of the environment is either learned or given.
% It is important to note that an RL algorithm can be either model-free or model-based depending on the type of the experience provided to the agent~\citep{sutton2018reinforcement}. 
Furthermore, a policy-based RL algorithm \citep{williams1992simple, schulman2017proximal, schulman2015trust} typically samples actions from a policy network $\policyAgent$ parameterized by neural networks with parameters $\RLParameters$. Given a state, $\policyAgent$ is trained to return the best action that maximizes the expected cumulative reward.

\section{The proposed method: \algName{}{}}\label{section_proposed_algorithm}

\begin{algorithm}[t]
   \caption{\algName{}. Best viewed in color. \textcolor{\rlCodeColor}{Typical RL code is colored in \rlCodeColor{}.}}
   \label{algorithm_pseudocode}
\begin{algorithmic}[1]
    \State {\bfseries Input:} the original user input $\userInput$, a predictive model $\PredModel$, a target class $\targetClass$, a reward function $\RewardFunc$, a state transition function $\StateTransFunc$, a proximity measure $\ProxmtDistance$, a proximity weight $\ProxmtLambda$, feature feasibility weights $\bm{\FeasbltWeights}$, maximum number of episodes $\MaxEpisodes$, maximum number of interventions per episode $\MaxInterventions$, discrete feature indicators $\discreteFeatures$, numbers of possible values of the discrete features $\{ \NumValuesDiscFeatIth{d} \vert d \in \discreteFeatures \}$.
    \State {\bfseries Optional Input:} non-actionable feature indicators $\nonActFeatures$, immutable feature indicators $\immuFeatures$, causal constraints $\CausalConstrants$, feature range constraints $\RangeConstrants$, in-distribution detector $\InDistribFunc$, a discount factor $\DiscountFactor$, a learning rate $\LearningRate$, a regularization weight $\WeightDecay$
    \State {\bfseries Output:} a CFE $\bm{O^{*}}$
    \State $\bm{O}=\{\emptyset\}$
    \State $E:=0$
    \While{$E < \MaxEpisodes$}
        \State \textcolor{\rlCodeColor}{$\bm{\tau}=\{\emptyset\}$ \textit{  \# Keep a record of (state, action, reward) pairs}} 
        \State $t:=0$
        \State $\bm{x_t}:=\userInput$
        \While{$t < \MaxInterventions$}
            \State $\action_t \sim \policy{\cdot}{\bm{x_t}}$ \textcolor{\codeCommentColor}{\textit{  \# Sample an action from the RL policy network}} \label{alg:policy}
            \State $\bm{x_{t+1}}:=\StateTransFunc(\bm{x_t},\action_t)$ \textcolor{\codeCommentColor}{\textit{  \# State transition from the current $\bm{x}$ to the next $\bm{x}$ }} \label{alg:state_transition}
            \State (Optionally, $\bm{x_{t+1}} := \RangeConstrants(\bm{x_{t+1}}) \cdot \CausalConstrants(\bm{x_{t+1}}) \cdot \immuFeatures(\bm{x_{t+1}})$) \label{alg:constraint}
            % update $\bm{x_{t+1}}$ according to $\RangeConstrants$, $\CausalConstrants$ and $\immuFeatures$) 
            \State $\Reward_{t+1} := \RewardFunc \left( \PredModel(\bm{x_{t+1}}), \targetClass, \ProxmtDistance(\userInput, \bm{x_{t+1}}, \bm{\FeasbltWeights}), \ProxmtLambda \right)$ \textcolor{\codeCommentColor}{\textit{  \# Compute the reward }} \label{alg:reward_proximity}
            \State \textcolor{\rlCodeColor}{$\bm{\tau} := \bm{\tau} \cup (\bm{x_t}, \action_t, \Reward_{t+1})$ \textit{  \# Add the pair to the record}}
            \If {$\PredModel(\bm{x_{t+1}}) = \targetClass$ and $\bm{x_{t+1}} \notin \bm{O}$} \textcolor{\codeCommentColor}{\textit{  \# If a new valid CFE is found }} \label{alg:prediction_equal_target}
                \State $\bm{O} := \bm{O} \cup \bm{x_{t+1}}$ (Optionally, if also $\InDistribFunc(\bm{x_{t+1}}) = \text{True}$) \label{alg:add_to_set}
                \State $t:=t+1$
                \State Break
            \EndIf
            \State $t:=t+1$
        \EndWhile
        \State $T := t$
        \For{$t = 0, 1, \dots, T-1$} \textcolor{\rlCodeColor}{\textit{ \# Update network parameters}}
            \State \textcolor{\rlCodeColor}{$\RLReturn := \sum_{t'=t+1}^{T} \DiscountFactor^{t'-t-1} \cdot \Reward_{t'}$}
            \State \textcolor{\rlCodeColor}{$\RLParameter := \RLParameter + \LearningRate \cdot \DiscountFactor^{t} \cdot \RLReturn \cdot \nabla \ln{\policy{\action_t}{\bm{x_t}}}$}
        \EndFor
        \State $E:=E+1$
    \EndWhile
    \State $\bm{O^{*}} := \min_{i} \ProxmtDistance(\userInput, \bm{O_{i}}, \bm{\FeasbltWeights})$ \textcolor{\codeCommentColor}{\textit{  \# Return the valid CFE with the lowest proximity }} \label{alg:min_proximity}
\end{algorithmic}
\end{algorithm}

In this work, we propose \fullAlgName{} (\algName{}), formulating CFE as an RL problem. In this setup, the RL environment is the CFE predictive model $\PredModel$. The RL state $\RLState$ is a sequence of CFEs beginning at the original user input $\userInput$ and ending in a final generated CFE $\bm{O^{*}}$. An action taken by the RL agent represents a small perturbation on the way from $\userInput$ to $\bm{O^{*}}$. A reward is a function of the predictive model and other objectives we introduce to maintain the properties discussed in Section~\ref{section_preliminaries} (more details below).

One-hot encoding is applied for categorical features. We assume that the prediction function of $\PredModel$ is computationally efficient, which is a common assumption in model-based RL \citep{sutton2018reinforcement}. 
% We also assume that continuous features in the original user input $\userInput$ are standardized to have mean $0$ and variance $1$.

%We propose \algName{}. 
\algName{} pseudocode is given in \Cref{algorithm_pseudocode}. Let $\userInput \in \mathbb{R}^{K \times D}$ denote a user input sample, where $K$ and $D$ denote the total number of time steps and features, respectively. To provide for plausibility, the $D$ features can optionally be categorized into actionable features $\actFeatures$, which the user can directly change; non-actionable features $\nonActFeatures$, which may change due to causal constraints but which the user cannot directly change \citep{sun2023cause}; and immutable features $\immuFeatures$, which may be used by the predictive model but which cannot change. $ \userInput$ is static if $K=1$ or temporal if $K>1$. The time complexity of the algorithm is $O(\MaxEpisodes \cdot \MaxInterventions)$.

\paragraph{Action (\Cref{alg:policy} of \Cref{algorithm_pseudocode})} \algName{} reduces the CFE search space from a multivariate time-series domain to a lower dimensional action space. Let $\policyAgent$ represent an RL policy network parameterized by neural networks with parameters $\RLParameters$. Each action $\action$ sampled from $\policyAgent$ is 3-dimensional $\action=\{\actionDimensionOne, \actionDimensionTwo, \actionDimensionThree\}$, where $\actionDimensionOne$ denotes the time step of the intervention, $\actionDimensionTwo$ denotes which feature to intervene on, and $\actionDimensionThree$ corresponds to the strength of the intervention.
% Let $\bm{P_{\cdot}(x)}$ denote one set of event probabilities in a categorical distribution that are non-negative and sum to $1$. 
% Let $\bm{\RLParameter_{1}(x)} = \bm{P_{1}(x)} \in \mathbb{R}^{K}$ and $\actionDimensionOne \sim \text{Cat}(K, \bm{\RLParameter_{1}(x)})$ denote which time step of $\RLState$ to intervene on. Let $\bm{\RLParameter_{2}(x)} = \bm{P_{2}(x)} \in \mathbb{R}^{D - |\nonActFeatures| - |\immuFeatures|}$ and $\actionDimensionTwo \sim \text{Cat}(D - |\nonActFeatures| - |\immuFeatures|, \bm{\RLParameter_{2}(x)})$ denote which feature of $\RLState$ to intervene on. 
% For each continuous feature $d \notin \nonActFeatures \cup \immuFeatures$, let $\RLParameter_{\{3, d\}}(x) = \mu_{\{3, d\}}(x) \in \mathbb{R}$ and $\RLParameter_{\{4, d\}}(x) = \sigma_{\{4, d\}}(x) \in \mathbb{R^{+}}$, which are the mean and standard deviation in a Gaussian distribution $N(\mu, \sigma^2)$. 
% For each discrete feature $d \notin \nonActFeatures \cup \immuFeatures$, let $\bm{\RLParameter_{\{5, d\}}(x)} = \bm{P_{\{5, d\}}(x)} \in \mathbb{R}^{\NumValuesDiscFeat^d}$. When $\actionDimensionTwo$ indicates a continuous feature, $\actionDimensionThree \sim N(\RLParameter_{\{3,\actionDimensionTwo\}}(\bm{x}), \RLParameter^{2}_{\{4,\actionDimensionTwo\}} (\bm{x}) )$ denotes how strong the intervention is for this continuous feature; when $\actionDimensionTwo$ indicates a discrete feature, $\actionDimensionThree \sim \text{Cat}(\NumValuesDiscFeat^{\actionDimensionTwo}, \bm{\RLParameter_{\{5, \actionDimensionTwo\}} (x)})$ denotes what the interventional value is for this discrete feature.
To be more specific, let ${DC}$ and ${DD}$ denote the numbers of actionable continuous and discrete features, respectively, where ${DC} + {DD} = |\actFeatures|$ and $|\actFeatures|$ denotes the total number of actionable features. Given an $\bm{x}$, the neural network parameterized by $\theta$ produces parameters to define four distributions $ \{ \actionDimensionOneParameters, \actionDimensionTwoParameters, \actionDimensionThreeParametersMean, \actionDimensionThreeParametersSTD, \actionDimensionThreeParametersDiscreteValue \} := \theta(\bm{x})$, such that $\actionDimensionOneParameters \in \mathbb{R}^{K}, \actionDimensionTwoParameters \in \mathbb{R}^{|\actFeatures|}, \actionDimensionThreeParametersMean \in \mathbb{R}^{DC}, \actionDimensionThreeParametersSTD \in \mathbb{R}^{DC} \text{ and } \actionDimensionThreeParametersDiscreteValue \in \mathbb{R}^{\NumValuesDiscFeat} \text{, where } \NumValuesDiscFeat = \sum_{i=1}^{DD} \NumValuesDiscFeatIth{i}$ and each $\bm{p}$ vector contains non-negative probabilities that sum to 1. The parameters $\actionDimensionOneParameters$ and $\actionDimensionTwoParameters$ define two categorical distributions from which $\actionDimensionOne$ and $\actionDimensionTwo$ are sampled, e.g. $\actionDimensionOne \sim \text{Cat}(\actionDimensionOneParameters)$ and $\actionDimensionTwo \sim \text{Cat}(\actionDimensionTwoParameters)$. When $\actionDimensionTwo$ is a continuous feature, the corresponding mean and standard deviation parameters $\actionDimensionThreeParameterMean{\actionDimensionTwo}$ and $\actionDimensionThreeParameterSTD{\actionDimensionTwo}$ define a normal distribution from which we sample how strong the intervention is for this continuous feature, e.g. $\actionDimensionThree \sim N(\actionDimensionThreeParameterMean{\actionDimensionTwo}, \actionDimensionThreeParameterSTD{\actionDimensionTwo}^2)$. When $\actionDimensionTwo$ is a discrete feature, the corresponding parameters $\actionDimensionThreeParameterDiscreteValue{\actionDimensionTwo}$ define a categorical distribution from which we sample what the interventional value is for this discrete feature, e.g. $\actionDimensionThree \sim \text{Cat}(\actionDimensionThreeParameterDiscreteValue{\actionDimensionTwo})$.

\paragraph{State Transition (\Cref{alg:state_transition} of \Cref{algorithm_pseudocode})} The state transition function $\StateTransFunc$ can be any appropriate function for an application domain. In \Cref{section_evaluation}, we define $\StateTransFunc(\bm{x_t},\action_t=\{\actionDimensionOne, \actionDimensionTwo, \actionDimensionThree\})$ as:
\begin{equation*}
\begin{aligned}
    x_{t+1}^{\{k, d\}} := 
\begin{cases}
    x_{t}^{\{k, d\}} + \actionDimensionThree & \text{for } k \geq \actionDimensionOne \text{ and } d=\actionDimensionTwo \text{ (when feature } d \text{ is continuous)} \\
    \actionDimensionThree & \text{for } k \geq \actionDimensionOne \text{ and } d=\actionDimensionTwo \text{ (when feature } d \text{ is discrete)} \\
    x_{t}^{\{k, d\}} & \text{otherwise}
\end{cases}
\end{aligned}
\end{equation*}
where $x_{t}^{\{k, d\}}$ denotes the $d$-th feature of the $t$-th $\bm{x}$ at the time step $k$.

\paragraph{Constraints (\Cref{alg:constraint} of \Cref{algorithm_pseudocode})} 
% Regarding causal constraints $\CausalConstrants$, many existing works require a complete causal graph or complete structural causal model (SCM) \citep{peters2017elements,karimi2020algorithmic,karimi2021algorithmic}. However, complete SCMs are often unavailable in practice \citep{survey_cfe_verma2020counterfactual,fastar_verma2022amortized}. \algName{} works with partial SCMs. 
% An (partial) SCM or $\RangeConstrants$ 
% After the state transition function $\StateTransFunc$ takes place, \algName{} checks for whether a new state violates any rules in $\CausalConstrants$ (or $\RangeConstrants$). If a rule is violated, \algName{} acts accordingly. 
% For example, it may choose to discard the change or set the corresponding value of the new state to a limiting value.
Constraints (i.e., $\RangeConstrants$ and $\CausalConstrants$) can be applied straightforwardly as sets of rules. For instance, a $\RangeConstrants$ constraint can be a rule that enforces that Feature 1 must remain within range $[-1, 1]$. If an action attempts to change the value of Feature 1 to $2$, the state transition can either be discarded or adjusted to cap the feature value at 1. Similarly, a $\CausalConstrants$ constraint such that Feature 1 is an interventional function of Feature 2, can be incorporated into the state transition function, ensuring that any action on Feature 2 updates Feature 1 accordingly. This dynamic is reflected in the reward function, which in turn influences the policy network, allowing the RL-based method to adapt accordingly.

\paragraph{Reward and Proximity (\Cref{alg:reward_proximity} of \Cref{algorithm_pseudocode}}) We define the reward function $\RewardFunc$ as:
\begin{equation*}
    \begin{aligned}
        \Reward := \RewardFunc \left( \PredModel(\bm{x}), \targetClass, \ProxmtDistance(\bm{\userInput}, \bm{x}, \bm{\FeasbltWeights}), \ProxmtLambda \right) = 
        \begin{cases}
            1 - \ProxmtLambda \cdot \ProxmtDistance(\bm{\userInput}, \bm{x}, \bm{\FeasbltWeights}) & \text{if } \PredModel(\bm{x}) = \targetClass \\
            0 & \text{otherwise}
        \end{cases}
    \end{aligned}
\end{equation*}
It combines a prediction reward ($1$ or $0$) and a weighted proximity loss $\ProxmtDistance$. $\ProxmtDistance$ is $0$ when $\PredModel(\bm{x}) \neq \targetClass$. Otherwise, in difficult settings where $\PredModel(\bm{x}) \neq \targetClass$ dominates over $\PredModel(\bm{x}) = \targetClass$, the RL agent would learn to produce CFEs that are too close to the original user input, which results in invalid CFEs. $\ProxmtLambda$ ensures that the reward is positive when $\PredModel(\bm{x}) = \targetClass$.

The proximity measure $\ProxmtDistance$ can be any suitable measures for the application domain. In \Cref{section_evaluation}, we define $\ProxmtDistance$ as the $L_1$-norm for continuous features and as the $L_0$-norm for discrete features, weighted by $\bm{\FeasbltWeights}$:
\begin{equation}
    \begin{aligned}
        \ProxmtDistance(\bm{x^d}, \bm{{x'}^d}, \FeasbltWeights^{d}) = 
        \begin{cases}
            \sum_{k=1}^{K} |x^{\{k,d\}} - {x'}^{\{k,d\}}| \cdot \FeasbltWeights^{d} \quad \text{ (if feature } d \text{ is continuous)} \\
            \sum_{k=1}^{K} I(x^{\{k,d\}} \neq {x'}^{\{k,d\}}) \cdot \FeasbltWeights^{d} \quad \text{ (if feature } d \text{ is discrete)} 
        \end{cases}
    \end{aligned}
    \label{eq:proximity}
\end{equation}
$\FeasbltWeights^{d}$ denotes the feasibility to change the $d$-th feature, which encodes the user's preference on altering this feature. $\bm{x}^{d}$ denotes the $d$-th feature of $\bm{x}$, and $x^{\{k, d\}}$ denotes the $d$-th feature of $\bm{x}$ at the time step $k$. \algName{} prefers to generate CFEs by altering features associated with small $\FeasbltWeights$. If a user does not specify preference on features, $\bm{\FeasbltWeights}$ is $1$ for all features. 

\paragraph{Classification and Regression (\Cref{alg:prediction_equal_target} of \Cref{algorithm_pseudocode})} \Cref{algorithm_pseudocode} describes \algName{} for a classification model $\PredModel$. As a model-agnostic method, \algName{} supports not only classification but also regression models. To work with a regression predictive model $\PredModel$, one can replace the first condition of \Cref{alg:prediction_equal_target} by $\targetClass_{\textit{lower}} \leq \PredModel(\bm{x_{t+1}}) \leq \targetClass_{\textit{upper}}$, where $\targetClass_{\textit{lower}}$ and $\targetClass_{\textit{upper}}$ represent the lower and upper bounds for the target regression value, respectively. 
% Although switching between classification and regression is simple with \algName{}, we focus on classification in \Cref{section_evaluation_quantitative}, because not all baselines support regression.

\paragraph{Output (\Cref{alg:add_to_set,alg:min_proximity} of \Cref{algorithm_pseudocode})} On \Cref{alg:add_to_set}, if a valid CFE is reached and is not already in the set $\bm{O}$, then it is added to $\bm{O}$. Optionally, an additional condition can be imposed such that a valid CFE is added to $\bm{O}$ only if it is also plausible.
Finally, on \Cref{alg:min_proximity}, \algName{} returns the CFE with the lowest proximity from the set $\bm{O}$.

\paragraph{Operating Without Training Datasets} The input $\userInput$ in Algorithm 1 is a testing sample $X_{\text{test}}$, i.e., the user sample to be explained or for which CFEs are generated. To clarify, the baseline methods require a training dataset $D_{\text{train}}$ to generate CFEs for $X_{\text{test}}$, whereas the proposed method operates directly on $X_{\text{test}}$ without the need for $D_{\text{train}}$. More details about how the baselines use $D_{\text{train}}$ is discussed in \Cref{section_related_works}.

\paragraph{Motivation of Using RL}
A search algorithm is required to solve the model-agnostic CFE problem without a training dataset. For multivariate time-series data, the CFE search space ($K \times D$) is large. BO-based CFE methods primarily focus on static or univariate time-series data \citep{spooner2021counterfactual,romashov2022baycon,huang2024tx} and do not scale well to multivariate time series. Optimization-based approaches, such as DiCE, are not model-agnostic since they require differentiable predictive models. Instead, we formulate the search problem as a continuous control task using reinforcement learning (RL), reducing it to a lower dimensional action space for improved scalability. Additionally, causal constraints are crucial for CFE methods \citep{survey_cfe_verma2020counterfactual}. While BO-based methods can impose simple value constraints (e.g., bounding ranges), they struggle with complex causal constraints. Our RL approach encodes causal constraints more naturally.

\subsection{Limitation}\label{section_limitation}
Features on different scales can impact the performance of models like neural networks. Without training datasets, standardizing the features becomes impractical. One approach is to leverage $\FeasbltWeights$ to mitigate the impact of continuous features on different scales. Assuming domain knowledge about the ranges of feature values (which is often available in practice), one can assign smaller $\FeasbltWeights$ to continuous features on larger scales and larger $\FeasbltWeights$ to continuous features on smaller scales. 
% This approach relaxes the standardization assumption. 
In our evaluation, we assume that the continuous features have a mean of 0 and a variance of 1. We leave the evaluation of this approach or a more sophisticated approach for future work.

The performance of the proposed model is unaffected by the size of the predictive model. Instead, its performance depends on the frequency with which the predictive model returns the desired prediction. We assume that the desired prediction returned by the predictive model is not sparse. This is a realistic assumption in many real-world domains. For example, numerous applicants with diverse personal characteristics applying for credit cards receive approval from the decision model of a bank. This assumption ensures that the RL agent receives enough reward signals to improve its performance. Without both training datasets and reward signals, it would be extremely challenging for ML methods to solve meaningful tasks.

\section{Evaluation}\label{section_evaluation}

In this section, we provide qualitative examples and quantitative experiment results to demonstrate the effectiveness of \algName{} for multivariate data-series data. The details about datasets and hyperparameters are provided in~\Cref{section_datasets} and \Cref{section_hyperparameters}, respectively.

\subsection{Qualitative Examples}\label{section_evaluation_qualitative}

We illustrate qualitative examples generated by \algName{} with two interpretable rule-based predictive models and an interpretable Life Expectancy dataset. Please refer to \Cref{section_prediction_models_rule_LE} for the definitions of the interpretable rule-based models. All examples in this section are generated using the first sample, which represents the country Albania, in the Life Expectancy dataset.

\subsubsection{Equal feature feasibility weights $\FeasbltWeights$}

\begin{figure*}[t]
     \centering
     \begin{subfigure}[t]{\textwidth}
         \centering
         \includegraphics[width=1.1\textwidth]{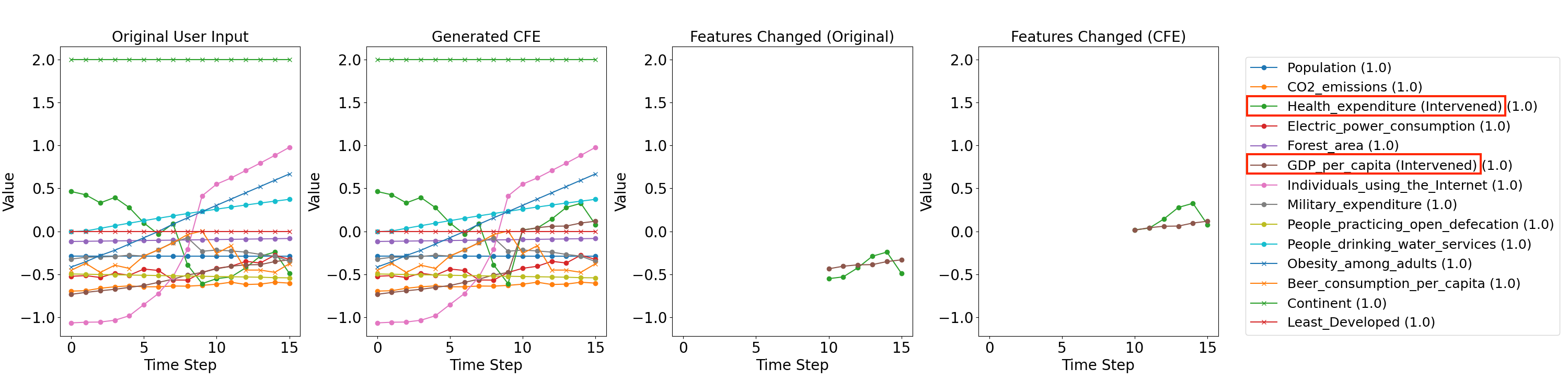}
         \caption{Rule-based model 1}
         \label{figure_qualitative_examples_weights_all_1_rule_1}
     \end{subfigure}
     \begin{subfigure}[t]{\textwidth}
         \centering
         \includegraphics[width=1.1\textwidth]{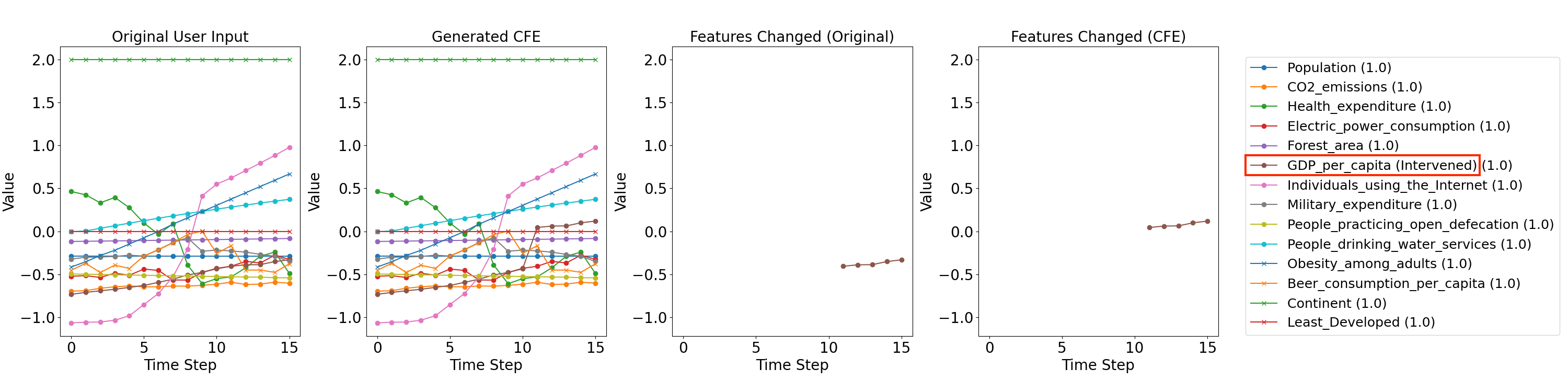}
         \caption{Rule-based model 2}
         \label{figure_qualitative_examples_weights_all_1_rule_2}
     \end{subfigure}        
     \caption{Qualitative examples with rule-based models and equal feature feasibility weights $\FeasbltWeights$. ``Original User Input'' shows the original input with the original feature values. ``Generated CFE'' shows the generated CFE with the modified features. The other two plots, ``Features Changed (Original)'' and ``Features Changed (CFE)'', omit most features and only show the modified features.}
     \label{figure_qualitative_examples_weights_all_1}
\end{figure*}

In \Cref{figure_qualitative_examples_weights_all_1}, all features have equal feasibility weights ($\FeasbltWeights=1.0$) as no user preferences are set. In \Cref{figure_qualitative_examples_weights_all_1_rule_1}, following the definition of rule-based model 1 (\Cref{section_prediction_models_rule_LE}), Albania's prediction is $0$ (i.e. undesired), because ``\textcolor{brown}{\textit{GDP-per-capita}}'' \textit{and} ``\textcolor{JungleGreen}{\textit{health-expenditure}}'' in the last 5 years are below 0. Accordingly, \algName{} generates a CFE by increasing these values above 0. In \Cref{figure_qualitative_examples_weights_all_1_rule_2}, if at least one of ``\textcolor{brown}{\textit{GDP-per-capita}}'' \textit{or} ``\textcolor{JungleGreen}{\textit{health-expenditure}}'' is above 0 in the last 5 years, then rule-based model 2 predicts 1 (i.e. desired). \algName{} generates a valid CFE for rule-based model 2 by raising ``\textcolor{brown}{\textit{GDP-per-capita}}'' above 0.

\subsubsection{Different feature feasibility weights $\FeasbltWeights$}

However, changing ``\textcolor{brown}{\textit{GDP-per-capita}}'' for Albania may be impractical. An alternative way to make rule-based model 2 predict 1 is to increase ``\textcolor{JungleGreen}{\textit{health-expenditure}}'' above 0. \algName{} can achieve this in three different ways: (1) marking ``\textcolor{brown}{\textit{GDP-per-capita}}'' as non-actionable; (2) assigning a small feasibility weight to ``\textcolor{JungleGreen}{\textit{health-expenditure}};'' or (3) assigning a large feasibility weight to ``\textcolor{brown}{\textit{GDP-per-capita}}.'' (1) is straightforward with \algName{}. Hence, we only present the results for (2) and (3). In Appendix \Cref{figure_qualitative_examples_different_feasibility_weights_health_0.1}, the feasibility weight $\FeasbltWeights$ for ``\textcolor{JungleGreen}{\textit{health-expenditure}}'' is set to $0.1$, ten times smaller than that of ``\textcolor{brown}{\textit{GDP-per-capita}},'' which remains unchanged as $1$. With the reduced feasibility weight for ``\textcolor{JungleGreen}{\textit{health-expenditure}},'' \algName{} generates a valid CFE by modifying ``\textcolor{JungleGreen}{\textit{health-expenditure}}.'' In Appendix \Cref{figure_qualitative_examples_different_feasibility_weights_GDP_10}, the feasibility weight for ``\textcolor{brown}{\textit{GDP-per-capita}}'' is set to $10$, ten times greater than other features. With the high feasibility weight for ``\textcolor{brown}{\textit{GDP-per-capita}},'' \algName{} preserves ``\textcolor{brown}{\textit{GDP-per-capita}}'' and looks for other features to achieve the desired prediction. As a result, \algName{} learns to alter ``\textcolor{JungleGreen}{\textit{health-expenditure}}''.

Next, we show that setting small feasibility weights on irrelevant features does not affect the CFEs generated by \algName{}. 
% when these features are not relevant to the model's prediction. 
In Appendix \Cref{figure_qualitative_examples_different_feasibility_weights_irrelavant}, although the feasibility weights for irrelevant features ``\textcolor{orange}{\textit{CO2-emissions}},'' ``\textcolor{magenta}{\textit{electric-power-consumption}},'' and ``\textcolor{Orchid}{\textit{forest-area}}'' are set to be ten times smaller than others, \algName{} still alters the relevant feature ``\textcolor{brown}{\textit{GDP-per-capita}}''.
% , which is the same as in \Cref{figure_qualitative_examples_weights_all_1_rule_2}. 
Additional results for different feasibility weights are provided in Appendix \Cref{figure_qualitative_examples_different_feasibility_weights_appendix}.

\begin{table*}[t]
\caption{Quantitative results with the eRing dataset for different predictive models and methods. Our proposed method \algName{} consistently achieves significantly better proximity and sparsity.}
% Note that success rate being equal to 0\% results in undefined validity rates because $N_{\textit{CFE}} = 0$.}
\begin{center}
\begin{tabular}{cccccccc}
\hline 
\multicolumn{1}{c}{\bf Predictive}  & 
\multicolumn{1}{c}{\bf $\NumInvalidSamples$}  & 
\multicolumn{1}{c}{\bf Methods}  & 
\multicolumn{1}{c}{\bf Success} & 
\multicolumn{1}{c}{\bf Validity} & 
\multicolumn{1}{c}{\bf Plausibility} &
\multicolumn{1}{c}{\bf Proximity} & 
\multicolumn{1}{c}{\bf Sparsity} \\
\multicolumn{1}{c}{\bf Model}  &  & 
\multicolumn{1}{c}{\bf}  & 
\multicolumn{1}{c}{\bf Rate (\%)} & 
\multicolumn{1}{c}{\bf Rate (\%)} & 
\multicolumn{1}{c}{\bf Rate (\%)} &
 & \\ 
 \hline 
\multirow{5}{*}{LSTM} & \multirow{5}{*}{14} & CoMTE & \textbf{100} & \textbf{100} & \textbf{100} & 346.746 & 260.0 \\
 &  & Native-Guide & \textbf{100} & \textbf{100} & \textbf{100} & 316.502 & 260.0 \\
 &  & CFRL & 0 & --- & --- & --- & --- \\
 &  & FastAR & 0 & --- & --- & --- & --- \\
 &  & \textbf{\algName{}} & \textbf{100} & \textbf{100} & \textbf{100} & \textbf{54.682} & \textbf{31.214} \\ 
 \hline 
\multirow{5}{*}{KNN} & \multirow{5}{*}{16} & CoMTE & \textbf{100} & \textbf{100} & \textbf{100} & 338.141 & 260.0 \\
 &  &Native-Guide & \textbf{100} & \textbf{100} & 93.75 & 
 $3.789 \times 10^{11}$ % 1378941378742.368 
 & 229.938 \\
&  &CFRL & \textbf{100} & \textbf{100} & \textbf{100} & 321.046 & 260.0 \\
&  &FastAR & 0 & --- & --- & --- & --- \\
&  & \textbf{\algName{}} & \textbf{100} & \textbf{100} & \textbf{100} & \textbf{144.937} & \textbf{86.688} \\ 
\hline 
\multirow{5}{1cm}{\centering Random Forest} & \multirow{5}{*}{15} & CoMTE  & \textbf{100} & \textbf{100} & \textbf{100} & 340.338 & 260.0 \\
 &  & Native-Guide & \textbf{100} & \textbf{100} & 93.333 & 
 $3.275 \times 10^{12}$ % 3274652404806.907 
 & 246.467 \\
&  & CFRL  & 0 & --- & --- & --- & --- \\
&  & FastAR & 0 & --- & --- & --- & --- \\
&  & \textbf{\algName{}} & \textbf{100} & \textbf{100} & \textbf{100} & \textbf{68.882} & \textbf{46.733} \\ 
\hline 
\multirow{5}{1cm}{\centering Rule Based 1} & \multirow{5}{*}{29} & CoMTE  & 0 & --- & --- & --- & --- \\
 &  & Native-Guide &  \textbf{100} & \textbf{100} & 44.828 & 
 $9.928 \times 10^{14}$ % 992843220439004.9 
 & 256.552 \\
&  & CFRL &  0 & --- & --- & --- & --- \\
&  & FastAR & 0 & --- & --- & --- & --- \\
&  & \textbf{\algName{}} & \textbf{100} & \textbf{100} & \textbf{100} & \textbf{103.953} & \textbf{63.345} \\ 
\hline 
\multirow{5}{1cm}{\centering Rule Based 2} & \multirow{5}{*}{25} & CoMTE & \textbf{100} & \textbf{100} & \textbf{100} & 347.295 & 260.0 \\
&  &  Native-Guide & \textbf{100} & \textbf{100} & 88.0 & 
$5.075 \times 10^{13}$ % 50753072875752.73 
& 245.4 \\
&  &  CFRL & 0 & --- & --- & --- & --- \\
&  &  FastAR & 0 & --- & --- & --- & --- \\
&  & \textbf{\algName{}} & \textbf{100} & \textbf{100} & \textbf{100} & \textbf{31.301} & \textbf{14.76} \\ \hline 
\end{tabular}
\end{center}
\label{tab:quantitative_exp_ERing}
\end{table*}

\subsection{Quantitative Experiments}\label{section_evaluation_quantitative}

In this section, we compare \algName{} to four baseline methods in 45 experiments, which correspond to nine real-world datasets each evaluated with five predictive models. Additional experiments are provided and analyzed in the appendix.

\paragraph{Datasets} Nine real-world multivariate time-series datasets are used for evaluation (\Cref{section_datasets} for details).
% : Life Expectancy, PEMS-SF, NATOPS, Heartbeat, Racket Sports, Basic Motions, eRing, Japanese Vowels, and Libras.

\paragraph{Baseline Methods} We benchmark \algName{} against four model-agnostic baseline methods: CoMTE \citep{comte_ates2021counterfactual}, Native-Guide \citep{native_guide_delaney2021instance}, CFRL \citep{cfrl_samoilescu2021model}, and FastAR \citep{fastar_verma2022amortized}. Optimization-based methods \citep{dice_mothilal2020explaining,sulem2022diverse,hsieh2021dice4el} are excluded from the comparison, because our predictive models are not restricted to differentiable models. For Native-Guide, we follow the approach of \citet{bahri2022temporal} by concatenating multivariate time-series samples into univariate time-series samples. Unlike the proposed \algName{}, the baselines require training datasets, i.e. either $(\bm{X_{\textit{train}}},\bm{Y_{\textit{train}}})$ or $(\bm{X_{\textit{train}}})$. We omit comparisons with other methods and use these popular methods as the representative baselines.

\paragraph{Predictive Models} Five predictive models are employed per dataset: a long short-term memory (LSTM) neural network, a k-nearest neighbor (KNN), a random forest and two interpretable rule models; see \Cref{section_prediction_models} for details. Given nine datasets, there is a total of 45 predictive models.

\subsubsection{Results}

The methods are evaluated using five metrics. Let $\NumInvalidSamples$ denote the total number of invalid samples, i.e., those classified as the undesired class by a predictive model, $N_{\textit{inv\_val}}$ denote the number of invalid samples for which a CFE method generates valid CFEs, $N_{\textit{val}}$ denote the number of valid CFEs generated by a CFE method, $N_{\textit{CFE}}$ denote the number of CFEs generated by a CFE method, and $N_{\textit{plau\_val}}$ denote the number of plausible and valid CFEs generated by a CFE method. 
We set feature feasibility weights $\FeasbltWeights^{d}=1$ for all the features $d \in D$. 

\Cref{tab:quantitative_exp_ERing} shows the results with the eRing dataset as an example. The proposed \algName{} consistently has the lowest proximity and sparsity. The results for all other datasets are given in Appendix \Cref{tab:quantitative_exp_LifeExpectancy,tab:quantitative_exp_pemssf,tab:quantitative_exp_NATOPS,tab:quantitative_exp_Heartbeat,tab:quantitative_exp_RacketSports,tab:quantitative_exp_BasicMotions,tab:quantitative_exp_JapaneseVowels,tab:quantitative_exp_Libras}. In this section, we analyze and compare the complete results, including these in the appendix.

Results dimmed in \textcolor{\resultIgnoreColor}{\resultIgnoreColor} in the tables are skipped from analysis and comparison. For Plausibility Rate, Proximity and Sparsity, we only compare the methods under 100\% success rates.
The reason is that if a method always generates CFEs that are close to the original invalid user input $\userInput$, even though these CFEs may often be invalid due to their closeness to $\userInput$, the plausibility rate, proximity and sparsity will always be superior. In the extreme case, if a method always returns the original but invalid $\userInput$, it would achieve a perfect plausibility rate, proximity and sparsity, but at the same time fail completely as a CFE method in terms of success rate.

\paragraph{Success Rate: $\frac{N_{\textit{inv\_val}}}{\NumInvalidSamples}$} There are two scenarios for a CFE method to fail: 1) no valid CFEs are generated; 2) no CFEs (either valid or invalid) are generated. For RL-based baselines, CFRL and FastAR fail with a 0\% success rate in 30/45 and 34/40 cases (excluding the 5 cases where FastAR crashes due to memory usage on \Cref{tab:quantitative_exp_pemssf}), respectively. \algName{} outperforms CFRL in 31/45 cases, and is on par with CFRL in 10/45 cases. In contrast, \algName{} underperforms CFRL in 4/45 cases. \algName{} outperforms FastAR in all 40/40 cases. Compared to the other baselines, Native-Guide, CoMTE and \algName{} fail with a 0\% success rate in 0/40 (excluding the 5 cases where Native-Guide crashes due to memory usage on \Cref{tab:quantitative_exp_pemssf}), 3/45 and 1/45 cases, respectively. \algName{} outperforms Native-Guide in 8/40 cases, and is on par with it in 25/40 cases. \algName{} underperforms Native-Guide in 7/40 cases. However, the minimum success rate of Native-Guide is 30.855\%, which is better than that of \algName{} (0.68\%). \algName{} underperforms CoMTE in success rate. \algName{}  produces lower success rates than CoMTE in 15/45 cases, achieves the same success rates in 27/45 cases, and outperforms it in only 3/45 cases. More comparison with CoMTE is provided at the end of this subsection.

It is important to note that: 1) We provide training datasets to the baselines (because they require training datasets to operate), but not to \algName{}. This additional information provided only to the baselines gives them an advantage over \algName{}. Without training datasets, the methods stop working except \algName{}. 2) In Appendix \Cref{tab:quan_results_increase_maximum_numbers} we show that the success rate of \algName{} can be further improved, e.g. from 0.68\% to 76.87\%.

\paragraph{Validity Rate: $\frac{N_{\textit{val}}}{N_{\textit{CFE}}}$} Both \algName{} and CoMTE ensure perfect validity rates by design (i.e. 100\%); they either produce a valid CFE or do not produce a CFE at all. In contrast, the other three baselines may return invalid CFEs; therefore, their validity rates can be lower than 100\%. 
%Furthermore, there are 3 cases where CoMTE fails completely with a 0\% success rate (\Cref{tab:quantitative_exp_NATOPS,tab:quantitative_exp_ERing,tab:quantitative_exp_JapaneseVowels}), and 1 such case for \algName{} (\Cref{tab:quantitative_exp_pemssf}). This results in undefined validity rates because $N_{\textit{CFE}} = 0$. 
Therefore, \algName{} and CoMTE outperform other baselines in terms of validity rate. 
% (i.e., 100\% in 44/45 cases). 
% However, if there were cases where \algName{} fail with a 0\% success rate, the validity rate for \algName{} would also be undefined.
% success rate and validity rate are identical for the three other baselines, because they always return one CFE (either valid or invalid). 
% Therefore, $\NumInvalidSamples = N_{\textit{CFE}}$ and $N_{\textit{inv\_val}} = N_{\textit{val}}$. However, it directly shows that the baselines may produce invalid CFEs (when a value $< 100\%$ ).

\paragraph{Plausibility Rate: $\frac{N_{\textit{plau\_val}}}{N_{\textit{val}}}$} The comparisons to CFRL and FastAR are skipped because more than half of the experiments yield 0\% success rates, and therefore, undefined plausibility rates. \algName{} outperforms and is on par with Native-Guide in 14/24 and 1/24 cases, respectively. \algName{} underperforms Native-Guide in 9/24 cases. \algName{} is on par with CoMTE in 8/27 cases and underperforms it in 19/27 cases. In summary, in terms of plausibility rate, CoMTE outperforms the proposed \algName{}, and \algName{} outperforms Native-Guide. Again, the baselines have the advantage by utilizing additional training information that is not provided to \algName{}. 

Additionally, one can enforce plausibility in \algName{} (\Cref{alg:add_to_set} of \Cref{algorithm_pseudocode}). \algName{} achieves 100\% plausibility rates at the cost of lower success rates and higher proximity and sparsity. Please see \Cref{section_evaluation_quantitative_plausibility} for details.

\paragraph{Proximity and Sparsity} 
Proximity and Sparsity are defined as the $L_1$-norm or $L_0$-norm, respectively, of the difference between a CFE and the original $\userInput$ \citep{fastar_verma2022amortized,cfrl_samoilescu2021model}.
% Proximity is defined as the unweighted \Cref{eq:proximity}. Sparsity is defined as the (unweighted) $L_0$-norm of the difference between the CFE and the original $\userInput$ for both continuous and discrete features, which is equivalent to the (unweighted) second equation of \Cref{eq:proximity}. 
Due to the aforementioned reason, proximity and sparsity are computed only with valid CFEs. Therefore, comparison with FastAR is skipped. {\em \algName{} outperforms all baselines in proximity and sparsity in all cases.}
% : \algName{} outperforms CFRL in all 10 cases, Native-Guide in all 24 cases, and CoMTE in all 25 cases. 
{\em It also surpasses the baselines by a large margin}. 
For example, there are 4, 8 or 15 cases where the proximity of \algName{} is at least 20 times, 10 times or 5 times lower than that of all the baselines, respectively (e.g., 16.183 vs. 220.552 in Appendix \Cref{tab:quantitative_exp_RacketSports}). 
Similarly, there are 3, 5 or 21 cases where the sparsity of \algName{} is at least 50 times, 20 times or 10 times lower than that of all the baselines, respectively (e.g., 20.587 vs. 1224.0 in Appendix \Cref{tab:quantitative_exp_NATOPS}).

\paragraph{Comparison with RL-based methods.} {\em \algName{} outperforms the two RL-based baselines, CFRL and FastAR, in all the metrics}. CFRL and FastAR often fail to generate valid CFEs for complex multivariate time-series data. Please note that this is not a criticism of CFRL or FastAR, because they are not designed for multivariate time-series data.

\paragraph{Comparison with CoMTE} Although CoMTE outperforms \algName{} in success rate and in plausibility rate, it is important to highlight that: 1) {\em CoMTE requires a training dataset, while \algName{} does not}. The better performance of CoMTE over \algName{} comes at the cost of needing more information and reduced versatility in practical applications. 2) CoMTE relies on finding distractors correctly classified as the target class. \Cref{tab:quantitative_exp_NATOPS,tab:quantitative_exp_ERing,tab:quantitative_exp_JapaneseVowels} for rule-based model 1 show that 
% when there are no samples with target class labels, 
when the predictive models classify all training samples as the undesired class, 
CoMTE fails completely with a 0\% success rate. In contrast, \algName{} is more versatile and can operate in such difficult situations. 3) Appendix \Cref{tab:quan_results_increase_maximum_numbers} shows that the success rates of \algName{} can be improved by increasing the maximum number of episodes $\MaxEpisodes$ or the maximum number of interventions per episode $\MaxInterventions$, which correspond to more exhaustive RL search.

\section{Conclusion}\label{section_conclusion}

In this paper, we introduce \algName{}, a model-agnostic RL-based method that generates CFEs for static and multivariate time-series data. \algName{} operates without requiring a training dataset, is compatible with both classification and regression predictive models, handles continuous and discrete features, and offers functionalities such as feature feasibility (i.e. user preference), feature actionability and causal constraints. We illustrate the effectiveness of \algName{} through qualitative examples and benchmark it against four state-of-the-art methods using nine real-world multivariate time-series datasets. Our results consistently show that \algName{} produces CFEs with significantly better proximity and sparsity. Future research includes extending the work to large language models, examining more advanced RL algorithms to potentially improve performance, using $\FeasbltWeights$ for features on different scales, and exploring alternation solutions other than RL for multivariate time-series data without training datasets.

%%%%%%%%%%%%%%%%%%%%%%%%%%%%%%%%%%%%%%%%%%%%%%%%%%%%%%%%%%%%

\newpage
\bibliography{main}
\bibliographystyle{tmlr}

%%%%%%%%%%%%%%%%%%%%%%%%%%%%%%%%%%%%%%%%%%%%%%%%%%%%%%%%%%%%

\newpage
\appendix

\section{Datasets}\label{section_datasets}

Nine real-world multivariate time-series datasets are used for evaluation in \Cref{section_evaluation_quantitative}:

\paragraph[]{Life Expectancy \footnote{https://www.kaggle.com/datasets/vrec99/life-expectancy-2000-2015}} 
% The Life Expectancy dataset has 119 samples. 
Each sample has 16 time steps (from 2000 to 2015) and 17 features per time step. All the features are interpretable. Please see~\Cref{tab:LE_features} for feature names and types. We remove ``\textit{Country Name}'' and ``\textit{Year}'' from the list of input features and use ``\textit{Life Expectancy}'' in 2015 as the label. Therefore, the dataset has $K=16$ and $D=14$ in our notation. We set $Y=1$ if ``\textit{Life Expectancy}'' in 2015 is greater or equal to 75 as the target class and otherwise $Y=0$ as the undesired class.

\paragraph[]{NATOPS \footnote{http://www.timeseriesclassification.com/description.php?Dataset=NATOPS}} The NATOPS dataset contains sensory data on hands, elbows, wrists and thumbs to classify movement types. 
% It has 180 samples. 
% Each sample has 51 time steps and 24 features per time step, i.e. 
$K=51$ and $D=24$. All the features in this dataset are continuous. There are 6 classes of different movements. We set class $4$ to $6$ as the target class.

\paragraph[]{PEMS-SF \footnote{https://www.timeseriesclassification.com/description.php?Dataset=PEMS-SF}} $K=144$ and $D=963$. The dataset contains transportation data from California. All the features are continuous. There are seven classes and we use the last four classes as the target class.

\paragraph[]{Heartbeat \footnote{http://www.timeseriesclassification.com/description.php?Dataset=Heartbeat}} 
% The Heartbeat dataset has 204 samples. 
% Each sample has 405 time steps and 61 features per time step, i.e. 
$K=405$ and $D=61$. All the features in this dataset are continuous. There are two classes: \textit{normal} heartbeat (target class) and \textit{abnormal} heartbeat (undesired class).

\paragraph[]{eRing\footnote{https://www.timeseriesclassification.com/description.php?Dataset=ERing}} 
% The eRing dataset has 30 samples. 
% Each sample has 65 time steps and 4 features per time step, i.e. 
$K=65$ and $D=4$. All the features in this dataset are continuous. There are six classes and we use the last three classes as the target class.

\paragraph[]{Racket Sports \footnote{https://www.timeseriesclassification.com/description.php?Dataset=RacketSports}} 
% The Racket Sports dataset has 151 samples. 
% Each sample has 30 time steps and 6 features per time step, i.e. 
$K=30$ and $D=6$. All the features in this dataset are continuous. There are four classes: ``\textit{Badminton Smash}'', ``\textit{Badminton Clear}'', ``\textit{Squash Forehand Boast}'' and ``\textit{Squash Backhand Boast}''. We use the last two classes as the target class.

\paragraph[]{Basic Motions \footnote{https://www.timeseriesclassification.com/description.php?Dataset=BasicMotions}} 
% The Basic Motions dataset has 40 samples. 
% Each sample has 100 time steps and 6 features per time step, i.e. 
$K=100$ and $D=6$. All the features in this dataset are continuous. There are four classes: ``\textit{Badminton}'', ``\textit{Running}'', ``\textit{Standing}'' and ``\textit{Walking}''. We use ``\textit{Standing}'' as the target class.

\paragraph[]{Japanese Vowels \footnote{https://www.timeseriesclassification.com/description.php?Dataset=JapaneseVowels}} 
% The Japanese Vowels dataset has 270 samples. 
% Each sample has 29 time steps and 12 features per time step, i.e. 
$K=29$ and $D=12$. All the features in this dataset are continuous. There are nine classes and we use the last five classes as the target class.

\paragraph[]{Libras \footnote{https://www.timeseriesclassification.com/description.php?Dataset=Libras}} 
% The Libras dataset has 180 samples. 
% Each sample has 45 time steps and 2 features per time step, i.e. 
$K=45$ and $D=2$. All the features in this dataset are continuous. There are 15 classes and we use the last eight classes as the target class.

All categorical features are one-hot encoded. All continuous features are standardized to have mean 0 and variance 1.

\begin{table*}[b]
\begin{center}
\caption{Features in the Life Expectancy dataset.}
\label{tab:LE_features}
\begin{tabular}{ll}
\multicolumn{1}{c}{\bf Features}  &\multicolumn{1}{c}{\bf Type} \\ 
\hline \\
\textit{Country Name} & Categorical \\
\textit{Year} & Categorical \\
\textit{Continent} & Categorical \\
\textit{Least Developed} & Categorical \\
\textit{Population} & Continuous \\
\textit{CO2 Emissions} & Continuous \\
\textit{Health Expenditure} & Continuous \\
\textit{Electric Power Consumption} & Continuous \\
\textit{Forest Area} & Continuous \\
\textit{GDP per Capita} & Continuous \\
\textit{Individuals Using the Internet} & Continuous \\
\textit{Military Expenditure} & Continuous \\
\textit{People Practicing Open Defecation} & Continuous \\
\textit{People Using at Least Basic Drinking Water Services} & Continuous \\
\textit{Obesity Among Adults} & Continuous \\
\textit{Beer Consumption per Capita} & Continuous \\
\hline \\
{\bf Label:} \textit{Life Expectancy} & Categorical \\
\end{tabular}
\end{center}
\end{table*}

\section{Predictive Models Used in \Cref{section_evaluation_quantitative}}\label{section_prediction_models}

Since \algName{} is model-agnostic, we expect it to work for predictive models with arbitrary hyperparameter values and architectures. In \Cref{section_evaluation_quantitative}, we evaluate the methods using five predictive models: two interpretable rule-
based models, a long short-term memory
(LSTM) neural network, a K-nearest neighbor (KNNs), and a random forest.

\subsection{Rule-based predictive models}\label{section_prediction_models_rule}

Two interpretable rule-based predictive models are implemented for each dataset.

\subsubsection*{Rule-based models for the Life Expectancy dataset}\label{section_prediction_models_rule_LE}

We employed two interpretable rule-based predictive models for the Life Expectancy dataset in \Cref{section_evaluation_qualitative} and \Cref{section_evaluation_quantitative}. 
Let $d_1$, $d_2$, $d_3$, $d_4$, $d_5$ denote the features ``\textit{least-developed}'', ``\textit{GDP-per-capita}'', ``\textit{health-expenditure}'', ``\textit{people-using-at-least-basic-drinking-water-services}'' and ``\textit{people-practicing-open-defecation}'', respectively. We define the first rule-based model as:
\begin{equation*}
    \begin{aligned}
        \PredModel(\bm{x}) = 
        \begin{cases}
            Y' & \text{ if } x^{\{k,d_1\}} = 0 \land x^{\{k,d_2\}} > 0 \land x^{\{k,d_3\}} > 0 \land x^{\{k,d_4\}} > 0 \land x^{\{k,d_5\}} < 0 \text{ for } K-4 \leq k \leq K \\
            Y & \text{ otherwise }
        \end{cases}
    \end{aligned}
\end{equation*}
and the second rule-based model as:
\begin{equation*}
    \begin{aligned}
        \PredModel(\bm{x}) = 
        \begin{cases}
            Y' & \text{ if } x^{\{k,d_1\}} = 0 \land \left( x^{\{k,d_2\}} > 0 \lor x^{\{k,d_3\}} > 0 \right) \land x^{\{k,d_4\}} > 0 \land x^{\{k,d_5\}} < 0 \text{ for } K-4 \leq k \leq K \\
            Y & \text{ otherwise }
        \end{cases}
    \end{aligned}
\end{equation*}
where $Y'$ is the target class and $Y$ is the undesired class.

\begin{figure*}[t]
     \centering
     \begin{subfigure}[t]{\textwidth}
         \centering
         \includegraphics[width=1.1\textwidth]{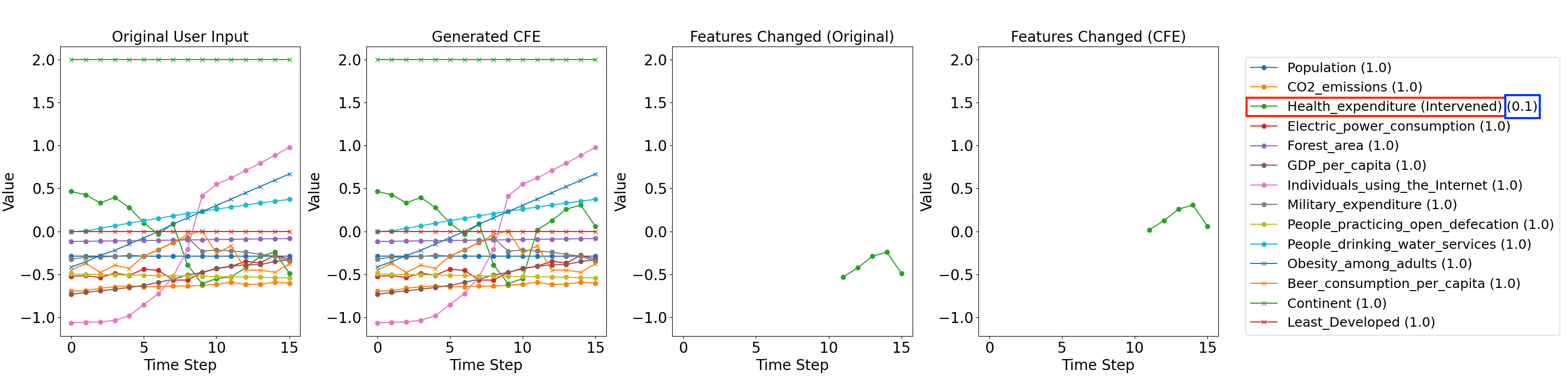}
         \caption{Rule-based model 2 Weight health 0.1}
         \label{figure_qualitative_examples_different_feasibility_weights_health_0.1}
     \end{subfigure}
     \begin{subfigure}[t]{\textwidth}
         \centering
         \includegraphics[width=1.1\textwidth]{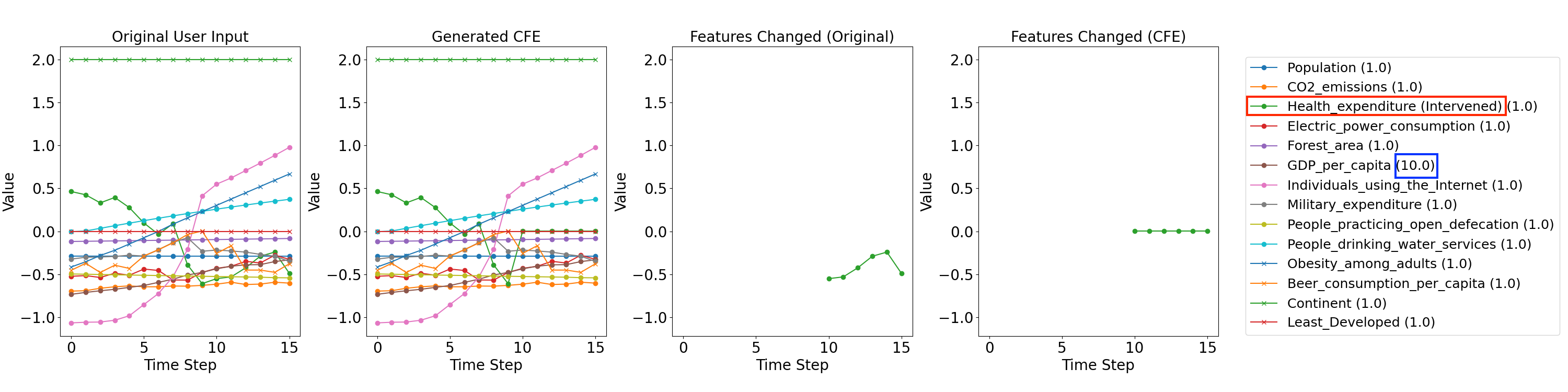}
         \caption{Rule-based model 2 Weight GDP 10}
         \label{figure_qualitative_examples_different_feasibility_weights_GDP_10}
     \end{subfigure}
     \begin{subfigure}[t]{\textwidth}
         \centering
         \includegraphics[width=1.1\textwidth]{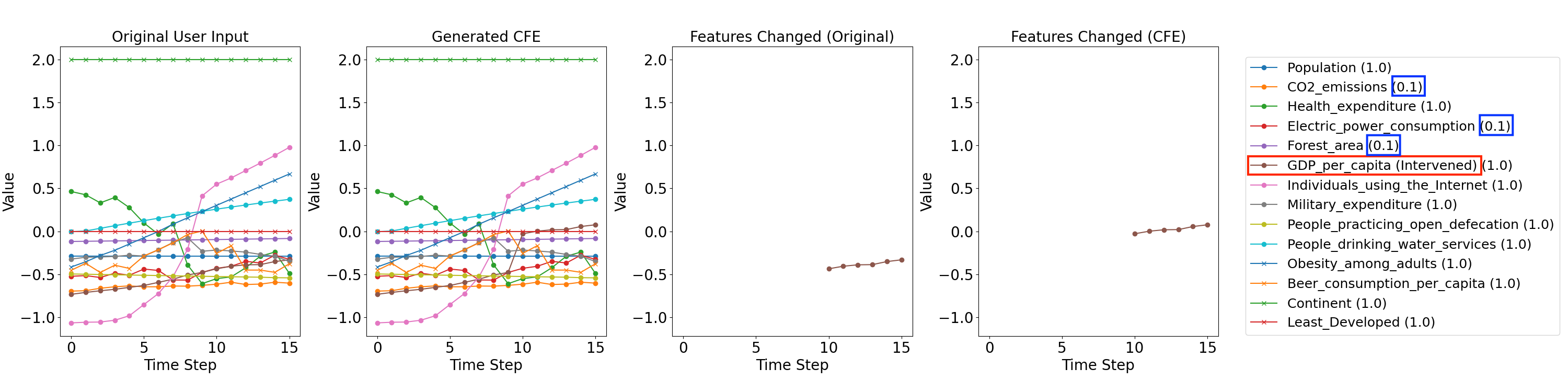}
         \caption{Rule-based model 2 Weight CO2 Elec Forest 0.1}
         \label{figure_qualitative_examples_different_feasibility_weights_irrelavant}
     \end{subfigure}
        \caption{Qualitative examples with rule-based model 2 and different feature feasibility weights $\FeasbltWeights$. }
        \label{figure_qualitative_examples_different_feasibility_weights}
\end{figure*}

\begin{figure*}[t]
     \centering
     \begin{subfigure}[t]{\textwidth}
         \centering
         \includegraphics[width=1.1\textwidth]{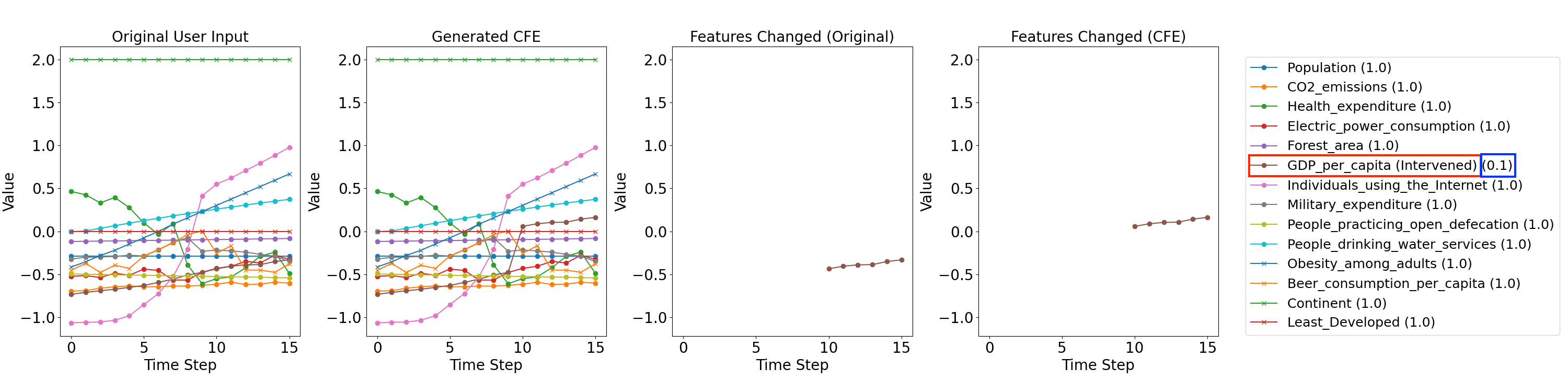}
         \caption{Rule-based model 2 Weight GDP 0.1}
     \end{subfigure}
     \begin{subfigure}[t]{\textwidth}
         \centering
         \includegraphics[width=1.1\textwidth]{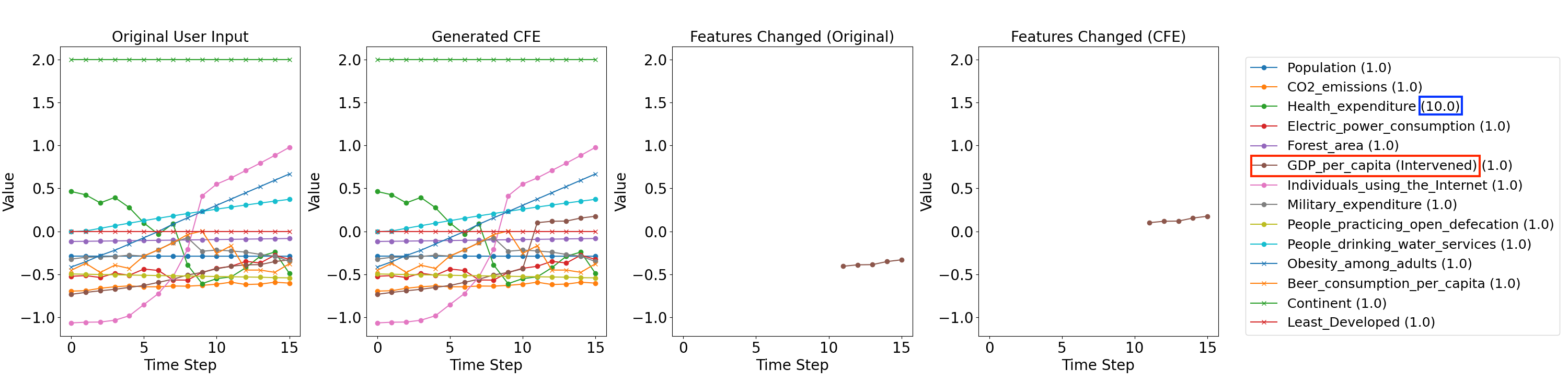}
         \caption{Rule-based model 2 Weight health 10}
     \end{subfigure}
        \caption{Qualitative examples with rule-based model 2 and different feature feasibility weights $\FeasbltWeights$. Among relevant features to the predictive model prediction, \algName{} prefers the features with smaller $\FeasbltWeights$.}
        \label{figure_qualitative_examples_different_feasibility_weights_appendix}
\end{figure*}

\subsubsection*{Rule-based models for the PEMS-SF dataset}\label{section_prediction_models_rule_pemssf}

Let $d_i$ denote the $i$-th feature. We define the first rule-based model as:\begin{equation*}
    \begin{aligned}
        \PredModel(\bm{x}) = 
        \begin{cases}
            Y' & \text{ if } x^{\{k,d_1\}} > 0 \land x^{\{k,d_{100}\}} > 0 \land x^{\{k,d_{300}\}} > 0 \text{ for } K-50 \leq k \leq K \\
            Y & \text{ otherwise }
        \end{cases}
    \end{aligned}
\end{equation*}
and the second rule-based model as:
\begin{equation*}
    \begin{aligned}
        \PredModel(\bm{x}) = 
        \begin{cases}
            Y' & \text{ if } x^{\{k,d_1\}} > 0 \lor x^{\{k,d_{100}\}} > 0 \lor x^{\{k,d_{300}\}} > 0 \text{ for } K-50 \leq k \leq K \\
            Y & \text{ otherwise }
        \end{cases}
    \end{aligned}
\end{equation*}
where $Y'$ is the target class and $Y$ is the undesired class.

\subsubsection*{Rule-based models for the NATOPS dataset}\label{section_prediction_models_rule_NATOPS}

Let $d_1$ and $d_2$ denote the features ``\textit{Hand tip left, X coordinate}'' and ``\textit{Hand tip right, X coordinate}'', respectively. We define the first rule-based model as:
\begin{equation*}
    \begin{aligned}
        \PredModel(\bm{x}) = 
        \begin{cases}
            Y' & \text{ if } x^{\{k,d_1\}} > 0 \land x^{\{k,d_2\}} > 0 \text{ for } K-9 \leq k \leq K \\
            Y & \text{ otherwise }
        \end{cases}
    \end{aligned}
\end{equation*}
and the second rule-based model as:
\begin{equation*}
    \begin{aligned}
        \PredModel(\bm{x}) = 
        \begin{cases}
            Y' & \text{ if } x^{\{k,d_1\}} > 0 \lor x^{\{k,d_2\}} > 0 \text{ for } K-9 \leq k \leq K \\
            Y & \text{ otherwise }
        \end{cases}
    \end{aligned}
\end{equation*}
where $Y'$ is one of the target classes and $Y$ is one of the undesired classes.

\subsubsection*{Rule-based models for the Heartbeat dataset}\label{section_prediction_models_rule_HB}

Let $d_i$ denote the $i$-th feature. We define the first rule-based model as:\begin{equation*}
    \begin{aligned}
        \PredModel(\bm{x}) = 
        \begin{cases}
            Y' & \text{ if } x^{\{k,d_1\}} > 0 \land x^{\{k,d_2\}} > 0 \land x^{\{k,d_3\}} > 0 \text{ for } K-4 \leq k \leq K \\
            Y & \text{ otherwise }
        \end{cases}
    \end{aligned}
\end{equation*}
and the second rule-based model as:
\begin{equation*}
    \begin{aligned}
        \PredModel(\bm{x}) = 
        \begin{cases}
            Y' & \text{ if } x^{\{k,d_1\}} > 0 \lor x^{\{k,d_2\}} > 0 \lor x^{\{k,d_3\}} > 0 \text{ for } K-4 \leq k \leq K \\
            Y & \text{ otherwise }
        \end{cases}
    \end{aligned}
\end{equation*}
where $Y'$ is the target class and $Y$ is the undesired class.

\subsubsection*{Rule-based models for the Racket Sports dataset}\label{section_prediction_models_rule_RS}

Let $d_i$ denote the $i$-th feature. We define the first rule-based model as:\begin{equation*}
    \begin{aligned}
        \PredModel(\bm{x}) = 
        \begin{cases}
            Y' & \text{ if } x^{\{k,d_1\}} > 0 \land x^{\{k,d_5\}} > 0 \text{ for } K-4 \leq k \leq K \\
            Y & \text{ otherwise }
        \end{cases}
    \end{aligned}
\end{equation*}
and the second rule-based model as:
\begin{equation*}
    \begin{aligned}
        \PredModel(\bm{x}) = 
        \begin{cases}
            Y' & \text{ if } x^{\{k,d_1\}} > 0 \lor x^{\{k,d_5\}} > 0 \text{ for } K-4 \leq k \leq K \\
            Y & \text{ otherwise }
        \end{cases}
    \end{aligned}
\end{equation*}
where $Y'$ is the target class and $Y$ is the undesired class.

\subsubsection*{Rule-based models for the Basic Motions dataset}\label{section_prediction_models_rule_BM}

Let $d_i$ denote the $i$-th feature. We define the first rule-based model as:\begin{equation*}
    \begin{aligned}
        \PredModel(\bm{x}) = 
        \begin{cases}
            Y' & \text{ if } x^{\{k,d_1\}} > 0 \land x^{\{k,d_3\}} > 0 \land x^{\{k,d_6\}} > 0 \text{ for } K-9 \leq k \leq K \\
            Y & \text{ otherwise }
        \end{cases}
    \end{aligned}
\end{equation*}
and the second rule-based model as:
\begin{equation*}
    \begin{aligned}
        \PredModel(\bm{x}) = 
        \begin{cases}
            Y' & \text{ if } x^{\{k,d_1\}} > 0 \lor x^{\{k,d_3\}} > 0 \lor x^{\{k,d_6\}} > 0 \text{ for } K-9 \leq k \leq K \\
            Y & \text{ otherwise }
        \end{cases}
    \end{aligned}
\end{equation*}
where $Y'$ is the target class ``\textit{Standing}'', and $Y$ is the undesired class.

\subsubsection*{Rule-based models for the eRing dataset}\label{section_prediction_models_rule_ering}

Let $d_i$ denote the $i$-th feature. We define the first rule-based model as:\begin{equation*}
    \begin{aligned}
        \PredModel(\bm{x}) = 
        \begin{cases}
            Y' & \text{ if } x^{\{k,d_2\}} > 0 \land x^{\{k,d_3\}} > 0 \text{ for } K-9 \leq k \leq K \\
            Y & \text{ otherwise }
        \end{cases}
    \end{aligned}
\end{equation*}
and the second rule-based model as:
\begin{equation*}
    \begin{aligned}
        \PredModel(\bm{x}) = 
        \begin{cases}
            Y' & \text{ if } x^{\{k,d_2\}} > 0 \lor x^{\{k,d_3\}} > 0 \text{ for } K-9 \leq k \leq K \\
            Y & \text{ otherwise }
        \end{cases}
    \end{aligned}
\end{equation*}
where $Y'$ is the target class and $Y$ is the undesired class.

\subsubsection*{Rule-based models for the Japanese Vowels dataset}\label{section_prediction_models_rule_JV}

Let $d_i$ denote the $i$-th feature. We define the first rule-based model as:\begin{equation*}
    \begin{aligned}
        \PredModel(\bm{x}) = 
        \begin{cases}
            Y' & \text{ if } x^{\{k,d_1\}} > 0 \land x^{\{k,d_6\}} > 0 \land x^{\{k,d_12\}} > 0 \text{ for } K-19 \leq k \leq K \\
            Y & \text{ otherwise }
        \end{cases}
    \end{aligned}
\end{equation*}
and the second rule-based model as:
\begin{equation*}
    \begin{aligned}
        \PredModel(\bm{x}) = 
        \begin{cases}
            Y' & \text{ if } x^{\{k,d_1\}} > 0 \lor x^{\{k,d_6\}} > 0 \lor x^{\{k,d_12\}} > 0 \text{ for } K-19 \leq k \leq K \\
            Y & \text{ otherwise }
        \end{cases}
    \end{aligned}
\end{equation*}
where $Y'$ is the target class and $Y$ is the undesired class.

\subsubsection*{Rule-based models for the Libras dataset}\label{section_prediction_models_rule_libras}

Let $d_i$ denote the $i$-th feature. We define the first rule-based model as:\begin{equation*}
    \begin{aligned}
        \PredModel(\bm{x}) = 
        \begin{cases}
            Y' & \text{ if } x^{\{k,d_1\}} > 0 \land x^{\{k,d_2\}} > 0 \text{ for } K-19 \leq k \leq K \\
            Y & \text{ otherwise }
        \end{cases}
    \end{aligned}
\end{equation*}
and the second rule-based model as:
\begin{equation*}
    \begin{aligned}
        \PredModel(\bm{x}) = 
        \begin{cases}
            Y' & \text{ if } x^{\{k,d_1\}} > 0 \lor x^{\{k,d_2\}} > 0 \text{ for } K-19 \leq k \leq K \\
            Y & \text{ otherwise }
        \end{cases}
    \end{aligned}
\end{equation*}
where $Y'$ is the target class and $Y$ is the undesired class.

\subsection{Other Predictive Models}

Besides the rule-based models, each of the following predictive models is also used for each dataset.

\paragraph{long short-term memory (LSTM).} The first layer of the neural network is a LSTM layer with 30 hidden states, followed by two linear layers. The first linear layer takes input of dimension of 30 and produces an output of dimension 60, then passes the output to a ReLU activation function. The second linear layer takes input of dimension of 60 and produces an output, then passes the output to a sigmoid activation function. We train the LSTM with learning rate 0.001 and weight decay 0.001 for 5000 epochs.

\paragraph{K-nearest neighbor (KNN).} The number of neighbors to use for prediction is $\sqrt{N}$, where $N$ denotes the number of samples in the dataset.

\paragraph{Random Forest.} The number of trees is $100$. The minimum number of samples required to split an internal node is 2. The minimum number of samples required to be at a leaf node is 1.

\section{Hyperparameters used in \Cref{section_evaluation}}\label{section_hyperparameters}

We use a unique set of hyperparameter values for \algName{} throughout the paper, unless otherwise stated, without fine-tuning them:
\begin{itemize}
    \item proximity weight $\ProxmtLambda=0.001$
    \item maximum number of interventions per episode $\MaxInterventions=100$
    \item maximum number of episodes $\MaxEpisodes=100$
    \item discount factor $\DiscountFactor=0.99$
    \item learning rate $\LearningRate=0.0001$
    \item regularization weight $\WeightDecay=0.0$
\end{itemize}

The RL policy network contains two hidden linear layers with 1000 and 100 neurons, respectively. Adam \citep{kingma2014adam} is used as the optimizer. 

It is important to note that \algName{} is not a supervised learning algorithm. As a reinforcement learning method, \algName{} does not rely on training datasets. Instead, it interacts with the environment (i.e., the predictive model $\PredModel$) for optimization. We select the set of hyperparameter values that works best for the Life Expectancy dataset from ten candidate sets of hyperparameter values. Although a more exhaustive hyperparameter search could potentially find another set of hyperparameters that produces better results, we leave this for future work.
% To increase the running speed of \algName{} in \Cref{section_evaluation_quantitative}, we decrease $\MaxEpisodes$ to $100$. As demonstrated in \Cref{tab:quan_results_increase_maximum_numbers} in the appendix, the success rate of \algName{} improves by increasing $\MaxEpisodes$ or $\MaxInterventions$. Therefore, we expect better results in \Cref{section_evaluation_quantitative} without decreasing $\MaxEpisodes$.
For the baseline methods, we use the hyperparameter values and architectures that are provided as default in their code~\footnote{The code is publicly available at: \newline CoMTE: https://github.com/peaclab/CoMTE \newline Native-Guide: https://github.com/e-delaney/Instance-Based\_CFE\_TSC \newline CFRL: https://docs.seldon.io/projects/alibi/en/stable/methods/CFRL.html \newline FastAR: https://github.com/vsahil/FastAR-RL-for-generating-AR}. All the experiments are conducted on CPU and with 32GB of RAM.

\section{Quantitative Experiments: \algName{} Versus $\mbox{\algName{}}_{\InDistrib}$}\label{section_evaluation_quantitative_plausibility}

In this section, we compare \algName{}, which does not enforce plausibility, with $\mbox{\algName{}}_{\InDistrib}$, which enforces plausibility.
For $\mbox{\algName{}}_{\InDistrib}$, $\InDistribFunc(\bm{x}) = \mbox{True}$ is applied (\Cref{alg:add_to_set} of \Cref{algorithm_pseudocode}) while any other hyperparameter values remain unchanged.
We employ local outlier factor (LOF) \citep{breunig2000lof} as the (optional) oracular in-distribution detector $\InDistribFunc$. LOF is a common method for assessing plausibility in CFE literature \citep{kanamori2020dace,native_guide_delaney2021instance,wang2021counterfactual,romashov2022baycon}. It employs KNNs to measure the degree to which a data point is unusual compared to others.

As shown in \Cref{tab:quantitative_exp_LifeExpectancy_plausibility,tab:quantitative_exp_NATOPS_plausibility,tab:quantitative_exp_Heartbeat_plausibility,tab:quantitative_exp_RacketSports_plausibility,tab:quantitative_exp_BasicMotions_plausibility,tab:quantitative_exp_ERing_plausibility,tab:quantitative_exp_JapaneseVowels_plausibility,tab:quantitative_exp_Libras_plausibility}, $\mbox{\algName{}}_{\InDistrib}$ achieves plausibility rates of 100\%. However, its success rates are lower than those of \algName{} in 22/39 cases. We further compare their proximity and sparsity under the same success rates. Although $\mbox{\algName{}}_{\InDistrib}$ gets higher proximity in 8/17 cases and higher sparsity in 7/17 cases, the changes in the values are small.

\begin{table*}[t]
\caption{Quantitative results with the Life Expectancy dataset.}
\label{tab:quantitative_exp_LifeExpectancy}
\begin{center}
\begin{tabular}{cccccccc}
\hline \\ 
\multicolumn{1}{c}{\bf Predictive}  & 
\multicolumn{1}{c}{\bf $\NumInvalidSamples$}  & 
\multicolumn{1}{c}{\bf Methods}  & 
\multicolumn{1}{c}{\bf Success} & 
\multicolumn{1}{c}{\bf Validity} & 
\multicolumn{1}{c}{\bf Plausibility} &
\multicolumn{1}{c}{\bf Proximity} & 
\multicolumn{1}{c}{\bf Sparsity} \\
\multicolumn{1}{c}{\bf Model}  &  
& 
\multicolumn{1}{c}{\bf}  & 
\multicolumn{1}{c}{\bf Rate (\%)} & 
\multicolumn{1}{c}{\bf Rate (\%)} & 
\multicolumn{1}{c}{\bf Rate (\%)} &
 & \\ 
\hline \\
\multirow{5}{*}{LSTM} & \multirow{5}{*}{63} & CoMTE & \textbf{100} & \textbf{100} & \textbf{100} & 37.475 & 201.54 \\
&& Native-Guide & \textbf{100} & \textbf{100} & 85.714 & \textbf{30.674} & \textbf{190.238} \\
&& CFRL & 0 & --- & --- & --- & --- \\
&& FastAR & 1.587 & 1.587 & \textcolor{\resultIgnoreColor}{100} & \textcolor{\resultIgnoreColor}{0.45} & \textcolor{\resultIgnoreColor}{1.0} \\
&&  \textbf{\algName{}} & 98.413 & \textbf{100} & \textcolor{\resultIgnoreColor}{80.645} & \textcolor{\resultIgnoreColor}{10.893} & \textcolor{\resultIgnoreColor}{64.065} \\
\hline \\
\multirow{5}{*}{KNN} & \multirow{5}{*}{68} & CoMTE & \textbf{100} & \textbf{100} & \textbf{100} & 46.366 & 204.176 \\
&& Native-Guide & \textbf{100} & \textbf{100} & 48.529 & 
$1.452 \times 10^{12} $ % 1451883310426.289 
& \textbf{200.574} \\
&& CFRL & \textbf{100} & \textbf{100} & \textbf{100} & \textbf{44.264} & 206.824 \\
&& FastAR & 0 & --- & --- & --- & --- \\
&& \textbf{\algName{}} & 85.294 & \textbf{100} & \textcolor{\resultIgnoreColor}{58.621} & \textcolor{\resultIgnoreColor}{19.756} & \textcolor{\resultIgnoreColor}{82.879} \\
\hline \\
\multirow{5}{1cm}{\centering Random Forest} & \multirow{5}{*}{62} & CoMTE & \textbf{100} & \textbf{100} & \textbf{100} & 33.752 & 199.468 \\
&&Native-Guide & \textbf{100} & \textbf{100} & 79.032 & 
$2.834 \times 10^{13}$ % 28343717711504.21 
& 200.548 \\
&&CFRL & \textbf{100} & \textbf{100} & \textbf{100} & 44.684 & 207.484 \\
&&FastAR & 0 & --- & --- & --- & --- \\
&&\textbf{\algName{}} & \textbf{100} & \textbf{100} & 96.774 & \textbf{8.724} & \textbf{49.661} \\
\hline \\
\multirow{5}{1cm}{\centering Rule-Based 1} & \multirow{5}{*}{87} & CoMTE & \textbf{100} & \textbf{100} & \textbf{100} & \textbf{47.542} & \textbf{203.747} \\
& & Native-Guide & 65.517 & 65.517 & \textcolor{\resultIgnoreColor}{66.667} & \textcolor{\resultIgnoreColor}{
$8.580 \times 10^{13}$ % 85800668810944.44
} & \textcolor{\resultIgnoreColor}{193.526} \\
& &  CFRL & 0 & --- & --- & --- & --- \\
& &  FastAR & 0 & --- & --- & --- & --- \\
& &  \textbf{\algName{}} & 82.759 & \textbf{100} & \textcolor{\resultIgnoreColor}{88.889} & \textcolor{\resultIgnoreColor}{10.566} & \textcolor{\resultIgnoreColor}{49.25} \\
\hline \\
\multirow{5}{1cm}{\centering Rule-Based 2} & \multirow{5}{*}{55} & CoMTE & \textbf{100} & \textbf{100} & \textbf{100} & \textbf{47.108} & \textbf{203.327} \\
&&Native-Guide & 72.727 & 72.727 & \textcolor{\resultIgnoreColor}{60.0} & \textcolor{\resultIgnoreColor}{
$7.749 \times 10^{13}$ % 77494497574951.52
} & \textcolor{\resultIgnoreColor}{183.025} \\
&&CFRL & 0 & --- & --- & --- & --- \\
&&FastAR & 0 & --- & --- & --- & --- \\
&&\textbf{\algName{}} & 81.818 & \textbf{100} & \textcolor{\resultIgnoreColor}{86.667} & \textcolor{\resultIgnoreColor}{11.238} & \textcolor{\resultIgnoreColor}{52.667} \\
\hline
\end{tabular}
\end{center}

\end{table*}

\begin{table*}[t]
\begin{center}
\caption{Quantitative results with the PEMS-SF dataset. Both FastAR and Native-Guide fail to run on this high-dimensional dataset, crashing due to memory issues, even after their memory allocation was increased to eight times that of the other methods.}
\label{tab:quantitative_exp_pemssf}
\begin{tabular}{cccccccc}
\hline \\ 
\multicolumn{1}{c}{\bf Predictive}  & 
\multicolumn{1}{c}{\bf $\NumInvalidSamples$}  & 
\multicolumn{1}{c}{\bf Methods}  & 
\multicolumn{1}{c}{\bf Success} & 
\multicolumn{1}{c}{\bf Validity} & 
\multicolumn{1}{c}{\bf Plausibility} &
\multicolumn{1}{c}{\bf Proximity} & 
\multicolumn{1}{c}{\bf Sparsity} \\
\multicolumn{1}{c}{\bf Model}  &  
& 
\multicolumn{1}{c}{\bf}  & 
\multicolumn{1}{c}{\bf Rate (\%)} & 
\multicolumn{1}{c}{\bf Rate (\%)} & 
\multicolumn{1}{c}{\bf Rate (\%)} &
 & \\ 
\hline \\
&&CoMTE & \textbf{100} & \textbf{100} & \textbf{81.25} & 139133.019 & \textbf{138672.0} \\
&&Native-Guide & --- & --- & --- & --- & --- \\
LSTM & 64 & CFRL & \textbf{100} & \textbf{100} & 0 & \textbf{139107.344} & \textbf{138672.0} \\
&&FastAR & --- & --- & --- & --- & --- \\
&&\textbf{\algName{}} &  40.625 & \textbf{100} & \textcolor{\resultIgnoreColor}{80.769} & \textcolor{\resultIgnoreColor}{1511.488} & \textcolor{\resultIgnoreColor}{961.654} \\
\hline \\
&&CoMTE & \textbf{100} & \textbf{100} & \textbf{74.118} & 139101.786 & \textbf{138672.0} \\
&&Native-Guide & --- & --- & --- & --- & --- \\
KNN & 96 &CFRL & \textbf{100} & \textbf{100} & 0 & \textbf{139100.448} & \textbf{138672.0} \\
&&FastAR & --- & --- & --- & --- & --- \\
&&\textbf{\algName{}} & 5.208 & \textbf{100}  & \textcolor{\resultIgnoreColor}{60.0} & \textcolor{\resultIgnoreColor}{1657.268} & \textcolor{\resultIgnoreColor}{1267.8} \\
\hline \\
&&CoMTE & \textbf{100} & \textbf{100} & \textbf{80.328} & 139134.124 & \textbf{138672.0} \\
Random & & Native-Guide & --- & --- & --- & --- & --- \\
Forest & 81 &CFRL & \textbf{100} & \textbf{100} & 0 & \textbf{139105.953} & \textbf{138672.0} \\
&&FastAR & --- & --- & --- & --- & --- \\
&&\textbf{\algName{}} & 0 & --- & --- & --- & --- \\
\hline \\
&&CoMTE & \textbf{100} & \textbf{100} & \textbf{92.806} & \textbf{139122.140} & \textbf{138671.986} \\
Rule & & Native-Guide & --- & --- & --- & --- & --- \\
Based 1 & 139 & CFRL & 0 & --- & --- & --- & --- \\
&&FastAR & --- & --- & --- & --- & --- \\
&&\textbf{\algName{}} & 96.403 & \textbf{100} & \textcolor{\resultIgnoreColor}{73.881} & \textcolor{\resultIgnoreColor}{3695.285} & \textcolor{\resultIgnoreColor}{3520.687} \\
\hline \\
&&CoMTE & \textbf{100} & \textbf{100} & \textbf{87.5} & 139067.896 & 138672.0 \\
Rule & & Native-Guide & --- & --- & --- & --- & --- \\
Based 2 & 24 &CFRL & 0 & --- & --- & --- & --- \\
&&FastAR & --- & --- & --- & --- & --- \\
&&\textbf{\algName{}} & \textbf{100} & \textbf{100} & 58.333 & \textbf{2586.612} & \textbf{2357.750} \\
\hline
\end{tabular}
\end{center}
\end{table*}

\begin{table*}[t]
\caption{Quantitative results with the NATOPS dataset.}
\label{tab:quantitative_exp_NATOPS}
\begin{center}
\begin{tabular}{cccccccc}
\hline \\ 
\multicolumn{1}{c}{\bf Predictive}  & 
\multicolumn{1}{c}{\bf $\NumInvalidSamples$}  & 
\multicolumn{1}{c}{\bf Methods}  & 
\multicolumn{1}{c}{\bf Success} & 
\multicolumn{1}{c}{\bf Validity} & 
\multicolumn{1}{c}{\bf Plausibility} &
\multicolumn{1}{c}{\bf Proximity} & 
\multicolumn{1}{c}{\bf Sparsity} \\
\multicolumn{1}{c}{\bf Model}  &  
& 
\multicolumn{1}{c}{\bf}  & 
\multicolumn{1}{c}{\bf Rate (\%)} & 
\multicolumn{1}{c}{\bf Rate (\%)} & 
\multicolumn{1}{c}{\bf Rate (\%)} &
 & \\ 
\hline \\
\multirow{5}{*}{LSTM} & \multirow{5}{*}{90} & CoMTE & \textbf{100} & \textbf{100} & \textbf{100} & 1285.262 & 1224.0 \\
&& Native-Guide & \textbf{100} & \textbf{100} & 63.333 & 
$1.270 \times 10^{13}$ % 12699535493460.361 
& 1158.133 \\
&& CFRL & 0 & --- & --- & --- & --- \\
&& FastAR & 0 & --- & --- & --- & --- \\
&& \textbf{\algName{}} & \textbf{100} & \textbf{100} & 28.889 & \textbf{227.184} & \textbf{135.1} \\ 
\hline \\
\multirow{5}{*}{KNN} & \multirow{5}{*}{93} & CoMTE & \textbf{100} & \textbf{100} & \textbf{100} & \textbf{1284.259} & \textbf{1224.0} \\
&& Native-Guide & \textbf{100} & \textbf{100} & 55.914 & 
$6.332 \times 10^{13} $ % 63320524766775.734 
& 1213.172 \\
&& CFRL & 0 & --- & --- & --- & --- \\
&& FastAR & 0 & --- & --- & --- & --- \\
&& \textbf{\algName{}} & 6.452 & \textbf{100} & \textcolor{\resultIgnoreColor}{50.0} & \textcolor{\resultIgnoreColor}{588.817} & \textcolor{\resultIgnoreColor}{496.333} \\ 
\hline \\
\multirow{5}{1cm}{\centering Random Forest} & \multirow{5}{*}{90}&CoMTE & \textbf{100} & \textbf{100} & \textbf{100} & 1285.888 & 1224.0 \\
&& Native-Guide & \textbf{100} & \textbf{100} & 28.889 & 
$3.193 \times 10^{12} $ % 3192621228993.281 
& 927.722 \\
&& CFRL & \textbf{100} & \textbf{100} & 0 & 1277.502 & 1224.0 \\
&& FastAR & 0 & --- & --- & --- & --- \\
&& \textbf{\algName{}} & \textbf{100} & \textbf{100} & 45.556 & \textbf{228.323} & \textbf{157.733} \\
\hline \\
\multirow{5}{1cm}{\centering Rule-Based 1} & \multirow{5}{*}{178} & CoMTE & 0 & --- & --- & --- & --- \\
&& Native-Guide & 66.854 & 66.854 & \textcolor{\resultIgnoreColor}{17.647} & \textcolor{\resultIgnoreColor}{
$1.349 \times 10^{14}$ % 134886624123713.94
} & \textcolor{\resultIgnoreColor}{1207.563} \\
&& CFRL & 0 & --- & --- & --- & --- \\
&& FastAR & 0 & --- & --- & --- & --- \\
&& \textbf{\algName{}} & \textbf{96.629} & \textbf{100} & \textcolor{\resultIgnoreColor}{86.047} & \textcolor{\resultIgnoreColor}{188.263} & \textcolor{\resultIgnoreColor}{144.105} \\
\hline \\
\multirow{5}{1cm}{\centering Rule-Based 2} & \multirow{5}{*}{126} & CoMTE & \textbf{100} & \textbf{100} & \textbf{100} & 1294.181 & 1224.0 \\
&& Native-Guide & 93.651 & 93.651 & \textcolor{\resultIgnoreColor}{72.881} & \textcolor{\resultIgnoreColor}{
$1.445 \times 10^{14}$ % 144484808550533.7
} & \textcolor{\resultIgnoreColor}{1208.458} \\
&& CFRL & 0 & --- & --- & --- & --- \\
&& FastAR & 0 & --- & --- & --- & --- \\
&& \textbf{\algName{}} & \textbf{100} & \textbf{100} & \textbf{100} & \textbf{33.756} & \textbf{20.587} \\ \hline
\end{tabular}
\end{center}

\end{table*}

\begin{table*}[t]
\begin{center}
\caption{Quantitative results with the Heartbeat dataset.}
\label{tab:quantitative_exp_Heartbeat}
\begin{tabular}{cccccccc}
\hline \\ 
\multicolumn{1}{c}{\bf Predictive}  & 
\multicolumn{1}{c}{\bf $\NumInvalidSamples$}  & 
\multicolumn{1}{c}{\bf Methods}  & 
\multicolumn{1}{c}{\bf Success} & 
\multicolumn{1}{c}{\bf Validity} & 
\multicolumn{1}{c}{\bf Plausibility} &
\multicolumn{1}{c}{\bf Proximity} & 
\multicolumn{1}{c}{\bf Sparsity} \\
\multicolumn{1}{c}{\bf Model}  &  
& 
\multicolumn{1}{c}{\bf}  & 
\multicolumn{1}{c}{\bf Rate (\%)} & 
\multicolumn{1}{c}{\bf Rate (\%)} & 
\multicolumn{1}{c}{\bf Rate (\%)} &
 & 
\\ \hline \\
&&CoMTE & \textbf{100} & \textbf{100} & \textbf{100} & \textbf{621.692} & 609.926 \\
&&Native-Guide & \textbf{100} & \textbf{100} & 77.206 & 
$1.075 \times 10^{11}$ % 107480237256.711 
& \textbf{591.346} \\
LSTM & 136 & CFRL & 2.941 & 2.941 & \textcolor{\resultIgnoreColor}{100} & \textcolor{\resultIgnoreColor}{627.946} & \textcolor{\resultIgnoreColor}{610.0} \\
&&FastAR & 0 & --- & --- & --- & --- \\
&& \textbf{\algName{}} & 97.794 & \textbf{100} & \textcolor{\resultIgnoreColor}{88.722} & \textcolor{\resultIgnoreColor}{16.825} & \textcolor{\resultIgnoreColor}{12.12} \\
\hline \\
&& CoMTE & \textbf{100} & \textbf{100} & \textbf{100} & \textbf{626.788} & \textbf{609.948} \\
&&Native-Guide & 97.396 & 97.396 & \textcolor{\resultIgnoreColor}{60.963} & \textcolor{\resultIgnoreColor}{
$4.749 \times 10^{12} $ % 4748729357720.746
} & \textcolor{\resultIgnoreColor}{600.824} \\
KNN & 192 & CFRL & 0 & --- & --- & --- & --- \\
&&FastAR & 0 & --- & --- & --- & --- \\
&&\textbf{\algName{}} & 72.396 & \textbf{100} & \textcolor{\resultIgnoreColor}{30.935} & \textcolor{\resultIgnoreColor}{145.611} & \textcolor{\resultIgnoreColor}{132.288} \\
\hline \\
&&CoMTE & \textbf{100} & \textbf{100} & \textbf{100} & \textbf{622.636} & \textbf{609.864} \\
Random & & Native-Guide & 65.986 & 65.986 & \textcolor{\resultIgnoreColor}{79.381} & \textcolor{\resultIgnoreColor}{
$1.877 \times 10^{12} $ % 1876988887157.566
} & \textcolor{\resultIgnoreColor}{600.68} \\
Forest & 147 & CFRL & 0 & --- & --- & --- & --- \\
&&FastAR & 0 & --- & --- & --- & --- \\
&&\textbf{\algName{}} & 0.68 & \textbf{100} & \textcolor{\resultIgnoreColor}{0} & \textcolor{\resultIgnoreColor}{57.664} & \textcolor{\resultIgnoreColor}{48.0} \\
\hline \\
&& CoMTE & \textbf{100} & \textbf{100} & \textbf{100} & \textbf{624.442} & \textbf{609.883} \\
Rule & & Native-Guide & 99.415 & 99.415 & \textcolor{\resultIgnoreColor}{78.235} & \textcolor{\resultIgnoreColor}{
$5.192 \times 10^{12} $ % 5192329737368.606
} & \textcolor{\resultIgnoreColor}{571.535} \\
Based 1 & 171 & CFRL & 0 & --- & --- & --- & --- \\
&&FastAR & 0 & --- & --- & --- & --- \\
&&\textbf{\algName{}} & 70.175 & \textbf{100} & \textcolor{\resultIgnoreColor}{43.333} & \textcolor{\resultIgnoreColor}{173.931} & \textcolor{\resultIgnoreColor}{162.692} \\
\hline \\
&&CoMTE & \textbf{100} & \textbf{100} & \textbf{100} & 619.454 & 609.917 \\
Rule & & Native-Guide & \textbf{100} & \textbf{100} & 82.5 & 
$2.236 \times 10^{11} $ % 225633275976.764 
& 589.658 \\
Based 2 & 120 & CFRL & 0 & --- & --- & --- & --- \\
&&FastAR & 0 & --- & --- & --- & --- \\
&&\textbf{\algName{}} & \textbf{100} & \textbf{100} & 90.833 & \textbf{13.842} & \textbf{9.008} \\ \hline
\end{tabular}
\end{center}

\end{table*}

\begin{table*}[t]

\caption{Quantitative results with the Racket Sports dataset.}
\label{tab:quantitative_exp_RacketSports}
\begin{center}
\begin{tabular}{cccccccc}
\hline \\ 
\multicolumn{1}{c}{\bf Predictive}  & 
\multicolumn{1}{c}{\bf $\NumInvalidSamples$}  & 
\multicolumn{1}{c}{\bf Methods}  & 
\multicolumn{1}{c}{\bf Success} & 
\multicolumn{1}{c}{\bf Validity} & 
\multicolumn{1}{c}{\bf Plausibility} &
\multicolumn{1}{c}{\bf Proximity} & 
\multicolumn{1}{c}{\bf Sparsity} \\
\multicolumn{1}{c}{\bf Model}  &  
& 
\multicolumn{1}{c}{\bf}  & 
\multicolumn{1}{c}{\bf Rate (\%)} & 
\multicolumn{1}{c}{\bf Rate (\%)} & 
\multicolumn{1}{c}{\bf Rate (\%)} &
 & 
\\ \hline \\
&&CoMTE & \textbf{100} & \textbf{100} & \textbf{100} & 214.707 & 180.0 \\
&&Native-Guide & \textbf{100} & \textbf{100} & 75.641 & 202.626 & 161.59 \\
LSTM & 78 & CFRL & 0 & --- & --- & --- & --- \\
& & FastAR & 14.103 & 14.103 & \textcolor{\resultIgnoreColor}{100} & \textcolor{\resultIgnoreColor}{1.786} & \textcolor{\resultIgnoreColor}{1.091} \\
&&\textbf{\algName{}} & \textbf{100} & \textbf{100} & 98.718 & \textbf{16.734} & \textbf{8.295} \\
\hline \\
&&CoMTE & \textbf{100} & \textbf{100} & \textbf{100} & 217.73 & 180.0 \\
&&Native-Guide & \textbf{100} & \textbf{100} & 76.786 & 
$5.320 \times 10^{12}$ % 5319556459890.491 
& 172.723 \\
KNN & 112 & CFRL & 0 & --- & --- & --- & --- \\
&&FastAR & 0 & --- & --- & --- & --- \\
&&\textbf{\algName{}} & \textbf{100} & \textbf{100} & 66.964 & \textbf{54.974} & \textbf{29.366} \\
\hline \\
&&CoMTE & \textbf{100} & \textbf{100} & \textbf{100} & 214.105 & 180.0 \\
Random & & Native-Guide & 98.78 & 98.78 & \textcolor{\resultIgnoreColor}{77.778} & \textcolor{\resultIgnoreColor}{
$3.529 \times 10^{12} $ % 3528683282737.359
} & \textcolor{\resultIgnoreColor}{167.222} \\
Forest & 82 & CFRL & 0 & --- & --- & --- & --- \\
&&FastAR & 0 & --- & --- & --- & --- \\
&&\textbf{\algName{}} & \textbf{100} & \textbf{100} & 90.244 & \textbf{43.695} & \textbf{26.232} \\
\hline \\
&&CoMTE & \textbf{100} & \textbf{100} & \textbf{100} & \textbf{222.762} & \textbf{180.0} \\
Rule & & Native-Guide & 94.595 & 94.595 & \textcolor{\resultIgnoreColor}{67.619} & \textcolor{\resultIgnoreColor}{
$1.170 \times 10^{14} $ % 116958541727411.39
} & \textcolor{\resultIgnoreColor}{172.486} \\
Based 1 & 111 & CFRL & 0 & --- & --- & --- & --- \\
&&FastAR & 0 & --- & --- & --- & --- \\
&&\textbf{\algName{}} & 98.198 & \textbf{100} & \textcolor{\resultIgnoreColor}{93.578} & \textcolor{\resultIgnoreColor}{24.46} & \textcolor{\resultIgnoreColor}{13.385} \\
\hline \\
&&CoMTE & \textbf{100} & \textbf{100} & \textbf{100} & 220.552 & 180.0 \\
Rule & & Native-Guide & \textbf{100} & \textbf{100} & 75.0 & 228.065 & 170.125 \\
Based 2 & 16 & CFRL & 0 & --- & --- & --- & --- \\
&&FastAR & 0 & --- & --- & --- & --- \\
&&\textbf{\algName{}} & \textbf{100} & \textbf{100} & \textbf{100} & \textbf{16.183} & \textbf{9.438} \\
\hline
\end{tabular}
\end{center}
\end{table*}

\begin{table*}[t]
\begin{center}
\caption{Quantitative results with the Basic Motions dataset.}
\label{tab:quantitative_exp_BasicMotions}
\begin{tabular}{cccccccc}
\hline \\ 
\multicolumn{1}{c}{\bf Predictive}  & 
\multicolumn{1}{c}{\bf $\NumInvalidSamples$}  & 
\multicolumn{1}{c}{\bf Methods}  & 
\multicolumn{1}{c}{\bf Success} & 
\multicolumn{1}{c}{\bf Validity} & 
\multicolumn{1}{c}{\bf Plausibility} &
\multicolumn{1}{c}{\bf Proximity} & 
\multicolumn{1}{c}{\bf Sparsity} \\
\multicolumn{1}{c}{\bf Model}  &  
& 
\multicolumn{1}{c}{\bf}  & 
\multicolumn{1}{c}{\bf Rate (\%)} & 
\multicolumn{1}{c}{\bf Rate (\%)} & 
\multicolumn{1}{c}{\bf Rate (\%)} &
 & \\
\hline \\
&&CoMTE & \textbf{100} & \textbf{100} & \textbf{100} & 782.188 & 600.0 \\
&&Native-Guide & \textbf{100} & \textbf{100} & 85.714 & 699.261 & 527.429 \\
LSTM & 14 & CFRL & \textbf{100} & \textbf{100} & \textbf{100} & 802.065 & 600.0 \\
&&FastAR & 7.143 & 7.143 & \textcolor{\resultIgnoreColor}{100} & \textcolor{\resultIgnoreColor}{2.85} & \textcolor{\resultIgnoreColor}{1.0} \\
&&\textbf{\algName{}} & \textbf{100} & \textbf{100} & \textbf{100} & \textbf{87.828} & \textbf{38.429} \\
\hline \\
&&CoMTE & \textbf{100} & \textbf{100} & \textbf{100} & 761.556 & 600.0 \\
&&Native-Guide & \textbf{100} & \textbf{100} & 78.947 & 706.719 & 562.526 \\
KNN & 19 & CFRL & \textbf{100} & \textbf{100} & \textbf{100} & 795.277 & 600.0 \\
&&FastAR & 0 & --- & --- & --- & --- \\
&&\textbf{\algName{}} & \textbf{100} & \textbf{100} & 94.737 & \textbf{206.984} & \textbf{121.421} \\
\hline \\
&&CoMTE & \textbf{100} & \textbf{100} & \textbf{100} & 751.741 & 600.0 \\
Random & & Native-Guide & \textbf{100} & \textbf{100} & 80.0 & 721.01 & 574.9 \\
Forest & 20 & CFRL & \textbf{100} & \textbf{100} & \textbf{100} & 773.104 & 600.0 \\
&&FastAR & 0 & --- & --- & --- & --- \\
&&\textbf{\algName{}} & \textbf{100} & \textbf{100} & 50.0 & \textbf{305.677} & \textbf{182.25} \\
\hline \\
&&CoMTE & \textbf{100} & \textbf{100} & \textbf{100} & \textbf{896.874} & \textbf{600.0} \\
Rule & & Native-Guide & 97.143 & 97.143 & \textcolor{\resultIgnoreColor}{88.235} & \textcolor{\resultIgnoreColor}{804.033} & \textcolor{\resultIgnoreColor}{584.971} \\
Based 1 & 35 & CFRL & 0 & --- & --- & --- & --- \\
&&FastAR & 0 & --- & --- & --- & --- \\
&&\textbf{\algName{}} & 97.143 & \textbf{100} & \textcolor{\resultIgnoreColor}{73.529} & \textcolor{\resultIgnoreColor}{133.66} & \textcolor{\resultIgnoreColor}{80.559} \\
\hline \\
&& CoMTE & \textbf{100} & \textbf{100} & \textbf{100} & 703.79 & 600.0 \\
Rule & & Native-Guide & \textbf{100} & \textbf{100} & 75.0 & 687.975 & 562.0 \\
Based 2 & 8 & CFRL & 0 & --- & --- & --- & --- \\
&&FastAR & 0 & --- & --- & --- & --- \\
&&\textbf{\algName{}} & \textbf{100} & \textbf{100} & \textbf{100} & \textbf{44.01} & \textbf{19.25} \\ \hline
\end{tabular}
\end{center}

\end{table*}

\begin{table*}[t]

\begin{center}
\caption{Quantitative results with the Japanese Vowels dataset.}
\label{tab:quantitative_exp_JapaneseVowels}
\begin{tabular}{cccccccc}
\hline \\ 
\multicolumn{1}{c}{\bf Predictive}  & 
\multicolumn{1}{c}{\bf $\NumInvalidSamples$}  & 
\multicolumn{1}{c}{\bf Methods}  & 
\multicolumn{1}{c}{\bf Success} & 
\multicolumn{1}{c}{\bf Validity} & 
\multicolumn{1}{c}{\bf Plausibility} &
\multicolumn{1}{c}{\bf Proximity} & 
\multicolumn{1}{c}{\bf Sparsity} \\
\multicolumn{1}{c}{\bf Model}  &  
& 
\multicolumn{1}{c}{\bf}  & 
\multicolumn{1}{c}{\bf Rate (\%)} & 
\multicolumn{1}{c}{\bf Rate (\%)} & 
\multicolumn{1}{c}{\bf Rate (\%)} &
 & \\ 
\hline \\
&&CoMTE & \textbf{100} & \textbf{100} & \textbf{100} & 330.771 & 300.0 \\
&&Native-Guide & \textbf{100} & \textbf{100} & 88.43 & 
$3.725 \times 10^{12} $ % 3725364012171.676 
& 281.05 \\
LSTM & 121&CFRL & \textbf{100} & \textbf{100} & \textbf{100} & 335.712 & 300.0 \\
&&FastAR & 0 & --- & --- & --- & --- \\
&&\textbf{\algName{}} & \textbf{100} & \textbf{100} & 99.174 & \textbf{35.3} & \textbf{19.413} \\ 
\hline \\
&&CoMTE & \textbf{100} & \textbf{100} & \textbf{100} & 331.845 & 300.0 \\
&&Native-Guide & \textbf{100} & \textbf{100} & 85.484 & 
$3.574 \times 10^{12} $ % 3574040130641.538 
& 283.419 \\
KNN & 124 & CFRL & \textbf{100} & \textbf{100} & \textbf{100} & 334.799 & 300.0 \\
&&FastAR & 1.613 & 1.613 & \textcolor{\resultIgnoreColor}{100} & \textcolor{\resultIgnoreColor}{1.6} & \textcolor{\resultIgnoreColor}{1.0} \\
&&\textbf{\algName{}} & \textbf{100} & \textbf{100} & 71.774 & \textbf{104.147} & \textbf{62.073} \\ 
\hline \\
&&CoMTE & \textbf{100} & \textbf{100} & \textbf{100} & 331.467 & 300.0 \\
Random & & Native-Guide & \textbf{100} & \textbf{100} & 89.167 & 
$9.824 \times 10^{12} $ % 9824478167355.428 
& 290.233 \\
Forest & 120 & CFRL & \textbf{100} & \textbf{100} & \textbf{100} & 334.707 & 300.0 \\
& & FastAR & 0 & --- & --- & --- & --- \\
&&\textbf{\algName{}} & \textbf{100} & \textbf{100} & 90.0 & \textbf{83.856} & \textbf{52.55} \\ 
\hline \\
&&CoMTE & 0 & --- & --- & --- & --- \\
Rule & & Native-Guide & 30.855 & 30.855 & \textcolor{\resultIgnoreColor}{78.313} & \textcolor{\resultIgnoreColor}{
$3.032 \times 10^{13} $ % 30323606490058.98
} & \textcolor{\resultIgnoreColor}{300.0} \\
Based 1 & 269 & CFRL & 0 & --- & --- & --- & --- \\
&&FastAR & 0 & --- & --- & --- & --- \\
&&\textbf{\algName{}} & \textbf{53.903} & \textbf{100} & \textcolor{\resultIgnoreColor}{57.241} & \textcolor{\resultIgnoreColor}{166.07} & \textcolor{\resultIgnoreColor}{123.869} \\ 
\hline \\
&&CoMTE & \textbf{100} & \textbf{100} & \textbf{100} & 330.597 & 300.0 \\
Rule & & Native-Guide & 67.114 & 67.114 & \textcolor{\resultIgnoreColor}{77.0} & \textcolor{\resultIgnoreColor}{
$6.965 \times 10^{12} $ % 6965451146717.747
} & \textcolor{\resultIgnoreColor}{289.92} \\
Based 2 & 149 &CFRL & 0 & --- & --- & --- & --- \\
&&FastAR & 0 & --- & --- & --- & --- \\
&&\textbf{\algName{}} & \textbf{100} & \textbf{100} & 97.987 & \textbf{38.206} & \textbf{18.436} \\ 
\hline
\end{tabular}
\end{center}
\end{table*}

\begin{table*}[t]
\begin{center}
\caption{Quantitative results with the Libras dataset.}
\label{tab:quantitative_exp_Libras}
\begin{tabular}{cccccccc}
\hline \\ 
\multicolumn{1}{c}{\bf Predictive}  & 
\multicolumn{1}{c}{\bf $\NumInvalidSamples$}  & 
\multicolumn{1}{c}{\bf Methods}  & 
\multicolumn{1}{c}{\bf Success} & 
\multicolumn{1}{c}{\bf Validity} & 
\multicolumn{1}{c}{\bf Plausibility} &
\multicolumn{1}{c}{\bf Proximity} & 
\multicolumn{1}{c}{\bf Sparsity} \\
\multicolumn{1}{c}{\bf Model}  &  
& 
\multicolumn{1}{c}{\bf}  & 
\multicolumn{1}{c}{\bf Rate (\%)} & 
\multicolumn{1}{c}{\bf Rate (\%)} & 
\multicolumn{1}{c}{\bf Rate (\%)} &
 & \\ 
\hline \\
&&CoMTE & \textbf{100} & \textbf{100} & \textbf{100} & 120.329 & 89.494 \\
&&Native-Guide & \textbf{100} & \textbf{100} & 64.045 & 111.969 & 88.382 \\
LSTM & 89 & CFRL & 0 & --- & --- & --- & --- \\
&&FastAR & 59.551 & 59.551 & \textcolor{\resultIgnoreColor}{64.151} & \textcolor{\resultIgnoreColor}{2.037} & \textcolor{\resultIgnoreColor}{1.094} \\
&&\textbf{\algName{}} & \textbf{100} & \textbf{100} & 32.584 & \textbf{21.421} & \textbf{12.73} \\
\hline \\
&&CoMTE & \textbf{100} & \textbf{100} & \textbf{100} & 118.379 & 88.989 \\
&&Native-Guide & \textbf{100} & \textbf{100} & 51.685 & 126.318 & 89.213 \\
KNN &89&CFRL & \textbf{100} & \textbf{100} & \textbf{100} & 139.54 & 90.0 \\
&&FastAR & 1.124 & 1.124 & \textcolor{\resultIgnoreColor}{0} & \textcolor{\resultIgnoreColor}{4.4} & \textcolor{\resultIgnoreColor}{3.0} \\
&&\textbf{\algName{}} & \textbf{100} & \textbf{100} & 14.607 & \textbf{45.306} & \textbf{24.112} \\ 
\hline \\
&&CoMTE & \textbf{100} & \textbf{100} & \textbf{100} & 108.074 & 88.393 \\
Random & & Native-Guide & \textbf{100} & \textbf{100} & 75.0 & 111.086 & 87.143 \\
Forest &84 &CFRL & 0 & --- & --- & --- & --- \\
&&FastAR & 0 & --- & --- & --- & --- \\
&&\textbf{\algName{}} & \textbf{100} & \textbf{100} & 13.095 & \textbf{50.151} & \textbf{27.024} \\ 
\hline \\
&&CoMTE & \textbf{100} & \textbf{100} & \textbf{100} & 144.053 & 88.836 \\
Rule & & Native-Guide & \textbf{100} & \textbf{100} & 87.069 & 135.861 & 88.836 \\
Based 1 &116 &CFRL & 0 & --- & --- & --- & --- \\
&&FastAR & 0 & --- & --- & --- & --- \\
&&\textbf{\algName{}} & \textbf{100} & \textbf{100} & 6.034 & \textbf{74.67} & \textbf{39.647} \\
\hline \\
&&CoMTE & \textbf{100} & \textbf{100} & \textbf{100} & 131.461 & 90.0 \\
Rule & & Native-Guide & \textbf{100} & \textbf{100} & 98.113 & 135.119 & 90.0 \\
Based 2 &53&CFRL & 0 & --- & --- & --- & --- \\
&&FastAR & 0 & --- & --- & --- & --- \\
&&\textbf{\algName{}} & \textbf{100} & \textbf{100} & 20.755 & \textbf{45.037} & \textbf{23.962} \\ 
\hline
\end{tabular}
\end{center}
\end{table*}

\begin{table*}[t]
\caption{Compare \algName{} and $\mbox{\algName{}}_{\InDistrib}$ with the Life Expectancy dataset.}
\label{tab:quantitative_exp_LifeExpectancy_plausibility}
\begin{center}
\begin{tabular}{cccccccc}
\hline \\ 
\multicolumn{1}{c}{\bf Predictive}  & 
\multicolumn{1}{c}{\bf $\NumInvalidSamples$}  & 
\multicolumn{1}{c}{\bf Methods}  & 
\multicolumn{1}{c}{\bf Success} & 
\multicolumn{1}{c}{\bf Validity} & 
\multicolumn{1}{c}{\bf Plausibility} &
\multicolumn{1}{c}{\bf Proximity} & 
\multicolumn{1}{c}{\bf Sparsity} \\
\multicolumn{1}{c}{\bf Model}  &  
& 
\multicolumn{1}{c}{\bf}  & 
\multicolumn{1}{c}{\bf Rate (\%)} & 
\multicolumn{1}{c}{\bf Rate (\%)} & 
\multicolumn{1}{c}{\bf Rate (\%)} &
 & 
\\ \hline \\
\multirow{2}{*}{LSTM}& \multirow{2}{*}{63} & \algName{} & 98.413 & 100 & 80.645 & \textcolor{\resultIgnoreColor}{10.893} & \textcolor{\resultIgnoreColor}{64.065} \\
& &  $\mbox{\algName{}}_{\InDistrib}$ & 93.651 & 100 & 100 & \textcolor{\resultIgnoreColor}{12.546} & \textcolor{\resultIgnoreColor}{57.695}
\\ \hline \\
\multirow{2}{*}{KNN}& \multirow{2}{*}{68} & \algName{} & 85.294 & 100 & 58.621 & \textcolor{\resultIgnoreColor}{19.756} & \textcolor{\resultIgnoreColor}{82.879} \\
&&$\mbox{\algName{}}_{\InDistrib}$ & 79.412 & 100 & 100 & \textcolor{\resultIgnoreColor}{25.653} & \textcolor{\resultIgnoreColor}{82.241} 
\\ \hline \\
\multirow{2}{1.5cm}{\centering Random Forest}& \multirow{2}{*}{62} & \algName{} & 100 & 100 & 96.774 & 8.724 & 49.661 \\
&&$\mbox{\algName{}}_{\InDistrib}$ & 100 & 100 & 100 & 8.984 & 49.242 \\
\hline \\
\multirow{2}{1.5cm}{\centering Rule Based 1}& \multirow{2}{*}{87} & \algName{} & 82.759 & 100 & 88.889 & \textcolor{\resultIgnoreColor}{10.566} & \textcolor{\resultIgnoreColor}{49.25} \\
&&$\mbox{\algName{}}_{\InDistrib}$ & 79.31 & 100 & 100 & \textcolor{\resultIgnoreColor}{10.063} & \textcolor{\resultIgnoreColor}{46.014} 
\\ \hline \\
\multirow{2}{1.5cm}{\centering Rule Based 2}& \multirow{2}{*}{55} & \algName{} & 81.818 & 100 & 86.667 & \textcolor{\resultIgnoreColor}{11.238} & \textcolor{\resultIgnoreColor}{52.667} \\
&&$\mbox{\algName{}}_{\InDistrib}$ & 76.364 & 100 & 100 & \textcolor{\resultIgnoreColor}{9.329} & \textcolor{\resultIgnoreColor}{46.69} \\ \hline
\end{tabular}
\end{center}

\end{table*}

\begin{table*}[t]
\begin{center}
\caption{Compare \algName{} and $\mbox{\algName{}}_{\InDistrib}$ with the NATOPS dataset.}
\label{tab:quantitative_exp_NATOPS_plausibility}
\begin{tabular}{cccccccc}
\hline \\ 
\multicolumn{1}{c}{\bf Predictive}  & 
\multicolumn{1}{c}{\bf $\NumInvalidSamples$}  & 
\multicolumn{1}{c}{\bf Methods}  & 
\multicolumn{1}{c}{\bf Success} & 
\multicolumn{1}{c}{\bf Validity} & 
\multicolumn{1}{c}{\bf Plausibility} &
\multicolumn{1}{c}{\bf Proximity} & 
\multicolumn{1}{c}{\bf Sparsity} \\
\multicolumn{1}{c}{\bf Model}  &  
& 
\multicolumn{1}{c}{\bf}  & 
\multicolumn{1}{c}{\bf Rate (\%)} & 
\multicolumn{1}{c}{\bf Rate (\%)} & 
\multicolumn{1}{c}{\bf Rate (\%)} &
 & 
\\ \hline \\
\multirow{2}{*}{LSTM}&\multirow{2}{*}{90}&\algName{} & 100 & 100 & 28.889 & \textcolor{\resultIgnoreColor}{227.184} & \textcolor{\resultIgnoreColor}{135.1} \\
&&$\mbox{\algName{}}_{\InDistrib}$ & 40.0 & 100 & 100 & \textcolor{\resultIgnoreColor}{192.739} & \textcolor{\resultIgnoreColor}{125.5} 
\\ \hline \\
\multirow{2}{*}{KNN}&\multirow{2}{*}{93}& \algName{} & 6.452 & 100 & 50.0 & \textcolor{\resultIgnoreColor}{588.817} & \textcolor{\resultIgnoreColor}{496.333} \\
&&$\mbox{\algName{}}_{\InDistrib}$ & 3.226 & 100 & 100 & \textcolor{\resultIgnoreColor}{109.59} & \textcolor{\resultIgnoreColor}{67.0} 
\\ \hline \\
\multirow{2}{1.5cm}{\centering Random Forest}&\multirow{2}{*}{90}& \algName{} & 100 & 100 & 45.556 & \textcolor{\resultIgnoreColor}{228.323} & \textcolor{\resultIgnoreColor}{157.733} \\
&&$\mbox{\algName{}}_{\InDistrib}$ & 63.333 & 100 & 100 & \textcolor{\resultIgnoreColor}{212.14} & \textcolor{\resultIgnoreColor}{156.333} 
\\ \hline \\
\multirow{2}{1.5cm}{\centering Rule Based 1}&\multirow{2}{*}{178}& \algName{} & 96.629 & 100 & 86.047 & \textcolor{\resultIgnoreColor}{188.263} & \textcolor{\resultIgnoreColor}{144.105} \\
&&$\mbox{\algName{}}_{\InDistrib}$ & 90.449 & 100 & 100 & \textcolor{\resultIgnoreColor}{133.034} & \textcolor{\resultIgnoreColor}{95.646}
\\ \hline \\
\multirow{2}{1.5cm}{\centering Rule Based 2 }&\multirow{2}{*}{126}& \algName{} & 100 & 100 & 100 & 33.756 & 20.587 \\
&& $\mbox{\algName{}}_{\InDistrib}$ & 100 & 100 & 100 & 33.756 & 20.587 \\ 
\hline
\end{tabular}
\end{center}
\end{table*}

\begin{table*}[t]
\begin{center}
\caption{Compare \algName{} and $\mbox{\algName{}}_{\InDistrib}$ with the Heartbeat dataset.}
\label{tab:quantitative_exp_Heartbeat_plausibility}
\begin{tabular}{cccccccc}
\hline \\ 
\multicolumn{1}{c}{\bf Predictive}  & 
\multicolumn{1}{c}{\bf $\NumInvalidSamples$}  & 
\multicolumn{1}{c}{\bf Methods}  & 
\multicolumn{1}{c}{\bf Success} & 
\multicolumn{1}{c}{\bf Validity} & 
\multicolumn{1}{c}{\bf Plausibility} &
\multicolumn{1}{c}{\bf Proximity} & 
\multicolumn{1}{c}{\bf Sparsity} \\
\multicolumn{1}{c}{\bf Model}  &  
& 
\multicolumn{1}{c}{\bf}  & 
\multicolumn{1}{c}{\bf Rate (\%)} & 
\multicolumn{1}{c}{\bf Rate (\%)} & 
\multicolumn{1}{c}{\bf Rate (\%)} &
 & 
\\ \hline \\
\multirow{2}{*}{LSTM}&\multirow{2}{*}{136}& \algName{} & 97.794 & 100 & 88.722 & 16.825 & 12.12 \\
&& $\mbox{\algName{}}_{\InDistrib}$ & 97.794 & 100 & 100 & 19.104 & 14.714 
\\ \hline \\
\multirow{2}{*}{KNN}&\multirow{2}{*}{192}& \algName{} & 72.396 & 100 & 30.935 & \textcolor{\resultIgnoreColor}{145.611} & \textcolor{\resultIgnoreColor}{132.288} \\
&&$\mbox{\algName{}}_{\InDistrib}$ & 25.521 & 100 & 100 & \textcolor{\resultIgnoreColor}{87.665} & \textcolor{\resultIgnoreColor}{77.245} 
\\ \hline \\
\multirow{2}{1.5cm}{\centering Random Forest}&\multirow{2}{*}{147}& \algName{} & 0.68 & 100 & 0 & \textcolor{\resultIgnoreColor}{57.664} & \textcolor{\resultIgnoreColor}{48.0} \\
&&$\mbox{\algName{}}_{\InDistrib}$ & 0 & --- & --- & --- & --- 
\\ \hline \\
\multirow{2}{1.5cm}{\centering Rule Based 1 }&\multirow{2}{*}{171}& \algName{} & 70.175 & 100 & 43.333 & \textcolor{\resultIgnoreColor}{173.931} & \textcolor{\resultIgnoreColor}{162.692} \\
&& $\mbox{\algName{}}_{\InDistrib}$ & 30.994 & 100 & 100 & \textcolor{\resultIgnoreColor}{83.472} & \textcolor{\resultIgnoreColor}{75.906}
\\ \hline \\
\multirow{2}{1.5cm}{\centering Rule Based 2 }&\multirow{2}{*}{120}& \algName{} & 100 & 100 & 90.833 & \textcolor{\resultIgnoreColor}{13.842} & \textcolor{\resultIgnoreColor}{9.008} \\
&& $\mbox{\algName{}}_{\InDistrib}$ & 99.167 & 100 & 100 & \textcolor{\resultIgnoreColor}{14.205} & \textcolor{\resultIgnoreColor}{9.613} \\ 
\hline
\end{tabular}
\end{center}
\end{table*}

\begin{table*}[t]

\begin{center}

\caption{Compare \algName{} and $\mbox{\algName{}}_{\InDistrib}$ with the Racket Sports dataset.}
\label{tab:quantitative_exp_RacketSports_plausibility}
\begin{tabular}{cccccccc}
\hline \\ 
\multicolumn{1}{c}{\bf Predictive}  & 
\multicolumn{1}{c}{\bf $\NumInvalidSamples$}  & 
\multicolumn{1}{c}{\bf Methods}  & 
\multicolumn{1}{c}{\bf Success} & 
\multicolumn{1}{c}{\bf Validity} & 
\multicolumn{1}{c}{\bf Plausibility} &
\multicolumn{1}{c}{\bf Proximity} & 
\multicolumn{1}{c}{\bf Sparsity} \\
\multicolumn{1}{c}{\bf Model}  &  
& 
\multicolumn{1}{c}{\bf}  & 
\multicolumn{1}{c}{\bf Rate (\%)} & 
\multicolumn{1}{c}{\bf Rate (\%)} & 
\multicolumn{1}{c}{\bf Rate (\%)} &
 & 
\\ \hline \\
\multirow{2}{*}{LSTM}&\multirow{2}{*}{78}& \algName{} & 100 & 100 & 98.718 & 16.734 & 8.295 \\
&&$\mbox{\algName{}}_{\InDistrib}$ & 100 & 100 & 100 & 16.961 & 8.423
\\ \hline \\
\multirow{2}{*}{KNN}&\multirow{2}{*}{112}& \algName{} & 100 & 100 & 66.964 & 54.974 & 29.366 \\
&&$\mbox{\algName{}}_{\InDistrib}$ & 100 & 100 & 100 & 64.028 & 36.42 
\\ \hline \\
\multirow{2}{1.5cm}{\centering Random Forest}&\multirow{2}{*}{82}& \algName{} & 100 & 100 & 90.244 & 43.695 & 26.232 \\
&& $\mbox{\algName{}}_{\InDistrib}$ & 100 & 100 & 100 & 44.611 & 27.768
\\ \hline \\
\multirow{2}{1.5cm}{\centering Rule Based 1 }&\multirow{2}{*}{111}& \algName{} & 98.198 & 100 & 93.578 & 24.46 & 13.385 \\
&& $\mbox{\algName{}}_{\InDistrib}$ & 98.198 & 100 & 100 & 25.226 & 13.633 
\\ \hline \\
 \multirow{2}{1.5cm}{\centering Rule Based 2}&\multirow{2}{*}{16} &\algName{} & 100 & 100 & 100 & 16.183 & 9.438 \\
&&$\mbox{\algName{}}_{\InDistrib}$ & 100 & 100 & 100 & 16.183 & 9.438 \\ \hline
\end{tabular}
\end{center}

\end{table*}

\begin{table*}[t]
\begin{center}
\caption{Compare \algName{} and $\mbox{\algName{}}_{\InDistrib}$ with the Basic Motions dataset.}
\label{tab:quantitative_exp_BasicMotions_plausibility}
\begin{tabular}{cccccccc}
\hline \\ 
\multicolumn{1}{c}{\bf Predictive}  & 
\multicolumn{1}{c}{\bf $\NumInvalidSamples$}  & 
\multicolumn{1}{c}{\bf Methods}  & 
\multicolumn{1}{c}{\bf Success} & 
\multicolumn{1}{c}{\bf Validity} & 
\multicolumn{1}{c}{\bf Plausibility} &
\multicolumn{1}{c}{\bf Proximity} & 
\multicolumn{1}{c}{\bf Sparsity} \\
\multicolumn{1}{c}{\bf Model}  &  
& 
\multicolumn{1}{c}{\bf}  & 
\multicolumn{1}{c}{\bf Rate (\%)} & 
\multicolumn{1}{c}{\bf Rate (\%)} & 
\multicolumn{1}{c}{\bf Rate (\%)} &
 & 
\\ \hline \\
\multirow{2}{*}{LSTM}&\multirow{2}{*}{14} & \algName{} & 100 & 100 & 100 & 87.828 & 38.429 \\
&&$\mbox{\algName{}}_{\InDistrib}$ & 100 & 100 & 100 & 87.828 & 38.429 
\\ \hline \\
\multirow{2}{*}{KNN}&\multirow{2}{*}{19} & \algName{} & 100 & 100 & 94.737 & 206.984 & 121.421 \\
&&$\mbox{\algName{}}_{\InDistrib}$ & 100 & 100 & 100 & 223.681 & 128.368
\\ \hline \\
\multirow{2}{1.5cm}{\centering Random Forest}&\multirow{2}{*}{20} & \algName{} & 100 & 100 & 50.0 & \textcolor{\resultIgnoreColor}{305.677} & \textcolor{\resultIgnoreColor}{182.25} \\
&&$\mbox{\algName{}}_{\InDistrib}$ & 95.0 & 100 & 100 & \textcolor{\resultIgnoreColor}{321.474} & \textcolor{\resultIgnoreColor}{199.895} 
\\ \hline \\
\multirow{2}{1.5cm}{\centering Rule Based 1}&\multirow{2}{*}{35} & \algName{} & 97.143 & 100 & 73.529 & 133.66 & 80.559 \\
&&$\mbox{\algName{}}_{\InDistrib}$ & 97.143 & 100 & 100 & 157.419 & 104.824
\\ \hline \\
\multirow{2}{1.5cm}{\centering Rule Based 2}&\multirow{2}{*}{8} & \algName{} & 100 & 100 & 100 & 44.01 & 19.25 \\
&&$\mbox{\algName{}}_{\InDistrib}$ & 100 & 100 & 100 & 44.01 & 19.25 \\ \hline
\end{tabular}
\end{center}
\end{table*}

\begin{table*}[t]
\begin{center}
\caption{Compare \algName{} and $\mbox{\algName{}}_{\InDistrib}$ with the eRing dataset.}
\label{tab:quantitative_exp_ERing_plausibility}
\begin{tabular}{cccccccc}
\hline \\ 
\multicolumn{1}{c}{\bf Predictive}  & 
\multicolumn{1}{c}{\bf $\NumInvalidSamples$}  & 
\multicolumn{1}{c}{\bf Methods}  & 
\multicolumn{1}{c}{\bf Success} & 
\multicolumn{1}{c}{\bf Validity} & 
\multicolumn{1}{c}{\bf Plausibility} &
\multicolumn{1}{c}{\bf Proximity} & 
\multicolumn{1}{c}{\bf Sparsity} \\
\multicolumn{1}{c}{\bf Model}  &  
& 
\multicolumn{1}{c}{\bf}  & 
\multicolumn{1}{c}{\bf Rate (\%)} & 
\multicolumn{1}{c}{\bf Rate (\%)} & 
\multicolumn{1}{c}{\bf Rate (\%)} &
 & 
\\ \hline \\
\multirow{2}{*}{LSTM}&\multirow{2}{*}{14} & \algName{} & 100 & 100 & 100 & 54.682 & 31.214 \\
&&$\mbox{\algName{}}_{\InDistrib}$ & 100 & 100 & 100 & 54.682 & 31.214
\\ \hline \\
\multirow{2}{*}{KNN}&\multirow{2}{*}{16} & \algName{} & 100 & 100 & 100 & 144.937 & 86.688 \\
&&$\mbox{\algName{}}_{\InDistrib}$ & 100 & 100 & 100 & 144.937 & 86.688 
\\ \hline \\
\multirow{2}{1.5cm}{\centering Random Forest}&\multirow{2}{*}{15} & \algName{} & 100 & 100 & 100 & 68.882 & 46.733 \\
&&$\mbox{\algName{}}_{\InDistrib}$ & 100 & 100 & 100 & 68.882 & 46.733 
\\ \hline \\
\multirow{2}{1.5cm}{\centering Rule Based 1}&\multirow{2}{*}{29} & \algName{} & 100 & 100 & 100 & 103.953 & 63.345 \\
&&$\mbox{\algName{}}_{\InDistrib}$ & 100 & 100 & 100 & 103.953 & 63.345 
\\ \hline \\
\multirow{2}{1.5cm}{\centering Rule Based 2}&\multirow{2}{*}{25} & \algName{} & 100 & 100 & 100 & 31.301 & 14.76 \\
&&$\mbox{\algName{}}_{\InDistrib}$ & 100 & 100 & 100 & 31.301 & 14.76 \\ \hline
\end{tabular}
\end{center}
\end{table*}

\begin{table*}[t]
\begin{center}
\caption{Compare \algName{} and $\mbox{\algName{}}_{\InDistrib}$ with the Japanese Vowels dataset.}
\label{tab:quantitative_exp_JapaneseVowels_plausibility}
\begin{tabular}{cccccccc}
\hline \\ 
\multicolumn{1}{c}{\bf Predictive}  & 
\multicolumn{1}{c}{\bf $\NumInvalidSamples$}  & 
\multicolumn{1}{c}{\bf Methods}  & 
\multicolumn{1}{c}{\bf Success} & 
\multicolumn{1}{c}{\bf Validity} & 
\multicolumn{1}{c}{\bf Plausibility} &
\multicolumn{1}{c}{\bf Proximity} & 
\multicolumn{1}{c}{\bf Sparsity} \\
\multicolumn{1}{c}{\bf Model}  &  
& 
\multicolumn{1}{c}{\bf}  & 
\multicolumn{1}{c}{\bf Rate (\%)} & 
\multicolumn{1}{c}{\bf Rate (\%)} & 
\multicolumn{1}{c}{\bf Rate (\%)} &
 & 
\\ \hline \\
\multirow{2}{*}{LSTM}&\multirow{2}{*}{121} & \algName{} & 100 & 100 & 99.174 & \textcolor{\resultIgnoreColor}{35.3} & \textcolor{\resultIgnoreColor}{19.413} \\
&&$\mbox{\algName{}}_{\InDistrib}$ & 99.174 & 100 & 100 & \textcolor{\resultIgnoreColor}{35.753} & \textcolor{\resultIgnoreColor}{19.833}
\\ \hline \\
\multirow{2}{*}{KNN}&\multirow{2}{*}{124} & \algName{} & 100 & 100 & 71.774 & \textcolor{\resultIgnoreColor}{104.147} & \textcolor{\resultIgnoreColor}{62.073} \\
&&$\mbox{\algName{}}_{\InDistrib}$ & 94.355 & 100 & 100 & \textcolor{\resultIgnoreColor}{114.437} & \textcolor{\resultIgnoreColor}{74.667} 
\\ \hline \\
\multirow{2}{1.5cm}{\centering Random Forest}&\multirow{2}{*}{120} & \algName{} & 100 & 100 & 90.0 & \textcolor{\resultIgnoreColor}{83.856} & \textcolor{\resultIgnoreColor}{52.55} \\
&&$\mbox{\algName{}}_{\InDistrib}$ & 98.333 & 100 & 100 & \textcolor{\resultIgnoreColor}{85.103} & \textcolor{\resultIgnoreColor}{54.314} 
\\ \hline \\
\multirow{2}{1.5cm}{\centering Rule Based 1}&\multirow{2}{*}{269} & \algName{} & 53.903 & 100 & 57.241 & \textcolor{\resultIgnoreColor}{166.07} & \textcolor{\resultIgnoreColor}{123.869} \\
&& $\mbox{\algName{}}_{\InDistrib}$ & 34.201 & 100 & 100 & \textcolor{\resultIgnoreColor}{135.475} & \textcolor{\resultIgnoreColor}{98.674} 
\\ \hline \\
\multirow{2}{1.5cm}{\centering Rule Based 2}&\multirow{2}{*}{149} & \algName{} & 100 & 100 & 97.987 & \textcolor{\resultIgnoreColor}{38.206} & \textcolor{\resultIgnoreColor}{18.436} \\
&& $\mbox{\algName{}}_{\InDistrib}$ & 99.329 & 100 & 100 & \textcolor{\resultIgnoreColor}{38.038} & \textcolor{\resultIgnoreColor}{18.926} \\ \hline
\end{tabular}
\end{center}

\end{table*}

\begin{table*}[t]
\begin{center}
\caption{Compare \algName{} and $\mbox{\algName{}}_{\InDistrib}$ with the Libras dataset.}
\label{tab:quantitative_exp_Libras_plausibility}
\begin{tabular}{cccccccc}
\hline \\ 
\multicolumn{1}{c}{\bf Predictive}  & 
\multicolumn{1}{c}{\bf $\NumInvalidSamples$}  & 
\multicolumn{1}{c}{\bf Methods}  & 
\multicolumn{1}{c}{\bf Success} & 
\multicolumn{1}{c}{\bf Validity} & 
\multicolumn{1}{c}{\bf Plausibility} &
\multicolumn{1}{c}{\bf Proximity} & 
\multicolumn{1}{c}{\bf Sparsity} \\
\multicolumn{1}{c}{\bf Model}  &  
& 
\multicolumn{1}{c}{\bf}  & 
\multicolumn{1}{c}{\bf Rate (\%)} & 
\multicolumn{1}{c}{\bf Rate (\%)} & 
\multicolumn{1}{c}{\bf Rate (\%)} &
 & 
\\ \hline \\
\multirow{2}{*}{LSTM}&\multirow{2}{*}{89} & \algName{} & 100 & 100 & 32.584 & \textcolor{\resultIgnoreColor}{21.421} & \textcolor{\resultIgnoreColor}{12.73} \\
&& $\mbox{\algName{}}_{\InDistrib}$ & 93.258 & 100 & 100 & \textcolor{\resultIgnoreColor}{39.937} & \textcolor{\resultIgnoreColor}{23.831}
\\ \hline \\
\multirow{2}{*}{KNN}&\multirow{2}{*}{89} & \algName{} & 100 & 100 & 14.607 & \textcolor{\resultIgnoreColor}{45.306} & \textcolor{\resultIgnoreColor}{24.112} \\
&& $\mbox{\algName{}}_{\InDistrib}$ & 88.764 & 100 & 100 & \textcolor{\resultIgnoreColor}{63.915} & \textcolor{\resultIgnoreColor}{37.215} 
\\ \hline \\
\multirow{2}{1.5cm}{\centering Random Forest}&\multirow{2}{*}{84} & \algName{} & 100 & 100 & 13.095 & \textcolor{\resultIgnoreColor}{50.151} & \textcolor{\resultIgnoreColor}{27.024} \\
&&$\mbox{\algName{}}_{\InDistrib}$ & 88.095 & 100 & 100 & \textcolor{\resultIgnoreColor}{61.098} & \textcolor{\resultIgnoreColor}{40.635} 
\\ \hline \\
\multirow{2}{1.5cm}{\centering Rule Based 1}&\multirow{2}{*}{116} & \algName{} & 100 & 100 & 6.034 & \textcolor{\resultIgnoreColor}{74.67} & \textcolor{\resultIgnoreColor}{39.647} \\
&&$\mbox{\algName{}}_{\InDistrib}$ & 50.862 & 100 & 100 & \textcolor{\resultIgnoreColor}{87.461} & \textcolor{\resultIgnoreColor}{50.695} \\
\hline \\
\multirow{2}{1.5cm}{\centering Rule Based 2}&\multirow{2}{*}{53} & \algName{} & 100 & 100 & 20.755 & \textcolor{\resultIgnoreColor}{45.037} & \textcolor{\resultIgnoreColor}{23.962} \\
&&$\mbox{\algName{}}_{\InDistrib}$ & 84.906 & 100 & 100 & \textcolor{\resultIgnoreColor}{75.344} & \textcolor{\resultIgnoreColor}{38.289} \\
\hline \\
\end{tabular}
\end{center}

\end{table*}

\begin{table*}[t]
\caption{This table shows that success rates are improved with increased maximum number of episodes $\MaxEpisodes$ and maximum number of interventions per episode $\MaxInterventions$, which correspond to more exhaustive RL search. We only test against the cases from \Cref{tab:quantitative_exp_LifeExpectancy,tab:quantitative_exp_pemssf,tab:quantitative_exp_NATOPS,tab:quantitative_exp_Heartbeat,tab:quantitative_exp_RacketSports,tab:quantitative_exp_BasicMotions,tab:quantitative_exp_ERing,tab:quantitative_exp_JapaneseVowels,tab:quantitative_exp_Libras} in which the default hyperparameters give success rates below 50\%.}
\label{tab:quan_results_increase_maximum_numbers}
\begin{subtable}{\textwidth}
\begin{center}
\begin{tabular}{llllll}
\multicolumn{1}{c}{\bf \algName{}}  & \multicolumn{1}{c}{\bf Success} & 
\multicolumn{1}{c}{\bf Validity} & 
\multicolumn{1}{c}{\bf Plausibility} &
\multicolumn{1}{c}{} & 
\multicolumn{1}{c}{} \\
\multicolumn{1}{c}{ \bf Hyperparameters}  & \multicolumn{1}{c}{\bf Rate (\%)} & 
\multicolumn{1}{c}{\bf Rate (\%)} & 
\multicolumn{1}{c}{\bf Rate (\%)} &
\multicolumn{1}{c}{\bf Proximity} & 
\multicolumn{1}{c}{\bf Sparsity} 
\\ \hline \\
 $\MaxEpisodes=100, \MaxInterventions=100$ & 0.68 & 100 & \textcolor{\resultIgnoreColor}{0} & \textcolor{\resultIgnoreColor}{57.664} & \textcolor{\resultIgnoreColor}{48.0} \\
$\MaxEpisodes=1000, \MaxInterventions=100$ & 10.204 & 100 & \textcolor{\resultIgnoreColor}{20.0} & \textcolor{\resultIgnoreColor}{267.045} & \textcolor{\resultIgnoreColor}{253.2} \\
$\MaxEpisodes=1000, \MaxInterventions=1000$ & 76.87 & 100 & \textcolor{\resultIgnoreColor}{4.425} & \textcolor{\resultIgnoreColor}{444.993} & \textcolor{\resultIgnoreColor}{417.336} \\
\end{tabular}
\end{center}
\caption{Dataset: Heartbeat. Predictive model: random forest.}
\label{tab:quan_results_HB_RF_maximum_numbers}
\end{subtable}

\begin{subtable}{\textwidth}
\begin{center}
\begin{tabular}{llllll}
\multicolumn{1}{c}{\bf \algName{}}  & \multicolumn{1}{c}{\bf Success} & 
\multicolumn{1}{c}{\bf Validity} & 
\multicolumn{1}{c}{\bf Plausibility} &
\multicolumn{1}{c}{} & 
\multicolumn{1}{c}{} \\
\multicolumn{1}{c}{\bf Hyperparameters}  & \multicolumn{1}{c}{\bf Rate (\%)} & 
\multicolumn{1}{c}{\bf Rate (\%)} & 
\multicolumn{1}{c}{\bf Rate (\%)} &
\multicolumn{1}{c}{\bf Proximity} & 
\multicolumn{1}{c}{\bf Sparsity} 
\\ \hline \\
$\MaxEpisodes=100, \MaxInterventions=100$ & 6.452 & 100 & \textcolor{\resultIgnoreColor}{50.0} & \textcolor{\resultIgnoreColor}{588.817} & \textcolor{\resultIgnoreColor}{496.333} \\
$\MaxEpisodes=1000, \MaxInterventions=100$ & 12.903 & 100 & \textcolor{\resultIgnoreColor}{33.333} & \textcolor{\resultIgnoreColor}{711.761} & \textcolor{\resultIgnoreColor}{595.583} \\
$\MaxEpisodes=10000, \MaxInterventions=100$ & 26.882 & 100 & \textcolor{\resultIgnoreColor}{16.0} & \textcolor{\resultIgnoreColor}{782.599} & \textcolor{\resultIgnoreColor}{627.96} \\
\end{tabular}
\end{center}
\caption{Dataset: NATOPS. Predictive model: KNN.}
\label{tab:quan_results_NATOPS_KNN_maximum_numbers}
\end{subtable}

% \begin{subtable}{\textwidth}
% \begin{center}
% \begin{tabular}{llllll}
% \multicolumn{1}{c}{\bf Methods}  & \multicolumn{1}{c}{\bf Success Rate} & 
% \multicolumn{1}{c}{\bf Validity Rate} & 
% \multicolumn{1}{c}{\bf Plausibility Rate} &
% \multicolumn{1}{c}{\bf Proximity} & 
% \multicolumn{1}{c}{\bf Sparsity} 
% \\ \hline \\
% \algName{} ($\MaxEpisodes=100, \MaxInterventions=100$)        & 72.396 & 100 & 30.935 & 145.611 & 132.288 \\
% \algName{} ($\MaxEpisodes=1000, \MaxInterventions=100$)        & 83.854 & 100 & 29.814 & 38.527 & 20.180 \\
% \algName{} ($\MaxEpisodes=1000, \MaxInterventions=1000$)       & 96.875 & 100 & 27.419 & 80.050 & 56.344 \\
% \end{tabular}
% \end{center}
% \caption{Dataset: Heartbeat. Predictive model: KNN.}
% \label{tab:quan_results_HB_KNN_maximum_numbers}
% \end{subtable}

\end{table*}

\end{document}